# A Semantically Enhanced Generative Foundation Model Improves Pathological Image Synthesis

## Author Information

Xianchao Guan[1,2,†], Zhiyuan Fan[1,†], Yifeng Wang[3,†], Fuqiang Chen[1,2], Yanjiang Zhou[1,2], Zengyang Che[1,2], Hongxue Meng[4], Xin Li[2], Yaowei Wang[1,2], Hongpeng Wang[1,2], Min Zhang[1], Heng Tao Shen[5], Zheng Zhang[1,2,*], and Yongbing Zhang[1,2,*]

[1] School of Computer Science and Technology, Harbin Institute of Technology (Shenzhen), Shenzhen, China

[2] Pengcheng Lab, Shenzhen, China

[3] School of Computer Science and Technology, Tsinghua University, Beijing, China

[4] Department of pathology, Harbin Medical University Cancer Hospital, Harbin, China

[5] School of Computer Science and Technology, Tongji University, Shanghai, China

[*] To whom correspondence should be addressed. Email: darrenzz219@gmail.com; ybzhang08@hit.edu.cn

[†] These authors contributed equally to this work.

# Abstract

The development of clinical-grade artificial intelligence in pathology is limited by the scarcity of diverse, high-quality annotated datasets. Generative models offer a potential solution but suffer from semantic instability and morphological hallucinations that compromise diagnostic reliability. To address this challenge, we introduce a Correlation-Regulated Alignment Framework for Tissue Synthesis (CRAFTS), the first generative foundation model for pathology-specific text-to-image synthesis. By leveraging a dual-stage training strategy on approximately 2.8 million image-caption pairs, CRAFTS incorporates a novel alignment mechanism that suppresses semantic drift to ensure biological accuracy. This model generates diverse pathological images spanning 30 cancer types, with quality rigorously validated by objective metrics and pathologist evaluations. Furthermore, CRAFTS-augmented datasets enhance the performance across various clinical tasks, including classification, cross-modal retrieval, self-supervised learning, and visual question answering. In addition, coupling CRAFTS with ControlNet enables precise control over tissue architecture from inputs such as nuclear segmentation masks and fluorescence images. By overcoming the critical barriers of data scarcity and privacy concerns, CRAFTS provides a limitless source of diverse, annotated histology data, effectively unlocking the creation of robust diagnostic tools for rare and complex cancer phenotypes.

# Introduction

Pathological images constitute the primary evidence base for clinical decision-making, enabling precise interpretation of tissue architecture and cellular morphology that underpins the detection, classification, and grading of diverse malignancies[1–3]. Their centrality to diagnostic practice has naturally positioned pathology at the forefront of computational medicine, where artificial intelligence (AI) systems increasingly rely on large-scale image corpora to learn the intricate visual signatures of disease[4–6]. Yet the rapid expansion of AI-driven pathology has sharply exposed a fundamental bottleneck: the limited availability of high-quality, expert-annotated images, particularly for rare, heterogeneous, or clinically nuanced cancer types[7–9]. Although whole-slide imaging technologies have transformed the digitization of pathology, current repositories remain fragmented in origin, uneven in quality, and insufficient in scope to support the training of scalable and robust foundation models[10–12]. This persistent data scarcity not only constrains algorithmic performance but also limits the deployment of AI systems across diverse clinical settings, underscoring an urgent need for more comprehensive and diverse pathological image resources[13,14].

Generative models offer a promising solution to this challenge by synthesizing realistic medical images that can augment training datasets[15,16,35,36]. Among the various generative approaches, Generative Adversarial Networks (GANs)[47] have been widely applied in medical imaging tasks[17,18,37,38]. GANs operate by learning the distribution of real images through two competing networks: a generator that creates synthetic images and a discriminator that evaluates their authenticity[19]. While GANs have demonstrated success in generating medical images, they often struggle to produce highly diverse, high-quality images, particularly in domains such as pathology, where fine-grained morphological features are critical[20,21]. These models are prone to mode collapse, in which they generate only a limited set of patterns, failing to capture the full diversity of disease manifestations[22,23,29,30]. More recently, diffusion models have emerged as a promising alternative to GANs, offering significant improvements in both diversity and quality[24,25,31,32]. Unlike GANs, diffusion models[48] iteratively denoise a random input, generating highly realistic images that better reflect the complex variations found in real medical datasets. These models have demonstrated success across a range of medical imaging applications, including the generation of chest radiographs, dermatological scans, and ultrasound images, providing realistic, clinically relevant data augmentation[15,26,28]. Their ability to generate high-quality, diverse medical images makes them an ideal choice for overcoming data scarcity, especially for rare diseases and underrepresented conditions[27,28,33,34].

Despite significant advances in generative models for the synthesis of pathology images, there are key limitations[20,26]. Some existing models condition on image features or gene expression data, overlooking the rich semantic context provided by textual descriptions. This omission prevents the models from capturing the full complexity of pathological conditions, particularly critical clinical details such as tumor heterogeneity, cellular architecture, and microenvironment interactions, which are essential for accurately representing the biological and clinical context of disease[27,29,30]. Textual descriptions offer valuable insights into these aspects of pathology, enabling a more nuanced understanding of disease mechanisms[39,40]. Therefore, integrating textual data as a conditioning signal is crucial for enhancing the clinical utility and accuracy of synthetic pathology images, enabling them to reflect a more comprehensive, contextually relevant depiction of disease.

In this study, we introduce a Correlation-Regulated Alignment Framework for Tissue Synthesis (CRAFTS), a groundbreaking generative foundation model specifically designed for pathology-focused text-to-image synthesis. CRAFTS addresses the challenge of limited high-quality annotated pathological image datasets, which have long been a critical barrier in computational pathology. Through large-scale training on millions of image-caption pairs, CRAFTS generates a highly diverse array of pathological images, covering 30 distinct cancer types and a broad spectrum of disease phenotypes. This generative capability is rigorously validated using both objective, quantitative metrics and expert evaluations from pathologists, ensuring that the synthetic images meet clinical standards of relevance and accuracy. In addition to its core image-generation capabilities, CRAFTS-generated synthetic data significantly enhances performance across various clinical tasks, including cancer classification, cross-modal retrieval, self-supervised learning, and visual question answering. Notably, CRAFTS also demonstrates its ability to precisely control image generation with few-shot, annotated images, offering promising opportunities to scale AI-driven diagnostic tools in oncology. This work represents a substantial advancement in pathological image synthesis, providing a scalable framework for generating clinically relevant data and promoting the development of AI-powered diagnostic tools. By addressing key data limitations, CRAFTS lays the foundation for the next generation of clinically applicable AI models in computational pathology, with the potential to revolutionize oncology diagnostics.

## Results

**Overview of CRAFTS**

CRAFTS is trained on a two-stage corpus that couples breadth with diagnostic fidelity. For pre-training, we curate ~1.2 million low-quality image-caption pairs from medical books and instructional videos[41–44], spanning wide stylistic variance and multi-scale pathological descriptions (Fig. 1a). Although visually noisy, these aligned pairs encode dense semantic co-occurrences of histomorphological concepts, enabling robust text-image grounding across diverse tissue contexts. We then use a high-quality fine-tuning set of ~1.6 million TCGA image-caption pairs[45], each annotated with one of 30 cancer types and paired with expert-level histological narratives (Fig. 1b). The TCGA-derived distribution provides clean morphological signals and explicit category priors, anchoring generation to real cancer-specific feature manifolds.

CRAFTS is instantiated on a latent diffusion backbone[46] tailored for pathology text-to-image synthesis (Fig. 1c). During pre-training, the model learns a coarse but scalable semantic correspondence between free-form pathological descriptions and visual latent codes, establishing cross-modal grounding under weakly paired data. Fine-tuning then injects explicit cancer-type priors into the generative process, using category-informed supervision to steer denoising trajectories toward type-specific histomorphological signatures and to disentangle subtle inter-class phenotypes. Throughout both stages, hallucination mitigation operates as a persistent training signal that penalizes semantic-image discordance and stabilizes text-image correlations, thereby enforcing faithful, clinically interpretable synthesis rather than visually plausible but semantically drifting outputs.

We evaluate CRAFTS with complementary objective metrics and blinded pathologist review, establishing high realism, semantic fidelity, and diversity (Fig. 1d). Synthetic data from CRAFTS consistently improves downstream performance, surpassing state-of-the-art augmentation and generative baselines in classification, cross-modal retrieval, self-supervised learning, and visual question answering (Fig. 1e), indicating that generated samples retain transferable diagnostic cues rather than superficial texture. Finally, by coupling CRAFTS with ControlNet[49], we enable few-shot controllable synthesis conditioned on structural prompts such as nuclear masks or fluorescence channels, yielding precise control over both tissue architecture and disease semantics (Fig. 1f). Collectively, CRAFTS establishes a scalable, semantically grounded, and controllable pathology synthesis paradigm for data-efficient computational oncology.

**Evaluation of synthetic images**

To comprehensively assess CRAFTS, we evaluate synthetic images along two complementary axes: visual fidelity, diversity, and text-image semantic concordance. Visual quality is examined through side-by-side qualitative inspection against real H&E fields and leading generative baselines, focusing on diagnostic microstructures and staining realism (Fig. 2a). In parallel, objective distributional and semantic metrics are computed using PLIP-based evaluations, including PLIP-FID[6] to quantify diversity and realism in feature space, PLIP-I[6] to measure similarity between synthetic and real images, and PLIP-T[6] to quantify how well generated images match their conditioning textual descriptions (Fig. 2b–d). Beyond automated scores, we perform blinded pathologist studies to gauge clinical plausibility and semantic faithfulness: experts (i) discriminate synthetic versus real slides to probe perceptual indistinguishability, and (ii) rank text-image alignment to assess whether generated morphology faithfully instantiates the prompt (Fig. 2e–f). Finally, we analyse cancer-type feature distributions of synthetic cohorts via embedding visualisation and silhouette-based separability, testing whether generated images preserve category-specific structure required for downstream learning (Fig. 2g).

**Qualitative comparison of synthetic images.** In the first two rows of Fig. 2a, where prompts describe pleomorphism, high nuclear-to-cytoplasmic ratios, disorganized architecture, and infiltrative growth, CRAFTS reproduces these cues with higher diagnostic precision than competing methods. Specifically, CRAFTS captures nuclear hyperchromasia and conspicuous nucleoli embedded within irregular, crowded epithelial nests, and reflects stromal invasion through disrupted gland-like organisation and dense cellularity. These features are only partially present or morphologically inconsistent in Stable Diffusion[46] and Imagen[51], and are often reduced to non-diagnostic textures in StyleGAN-T[52]. In the third and fourth rows, CRAFTS further demonstrates superior image quality: spindle-cell fascicles and collagenous stroma are rendered with coherent fibre orientation, crisp nuclear contours, and realistic H&E chromatin contrast; likewise, the pancreatic ductal adenocarcinoma example shows markedly clearer cytologic detail and structured desmoplastic background. By contrast, baselines exhibit frequent blur, washed or overly saturated eosin/basophilia, and unstable tissue architecture, limiting interpretability at the cellular level.

**Objective metric evaluation.** CRAFTS achieves a clear advantage in objective evaluations that probe both pathological realism and cross-modal faithfulness. First, in terms of visual quality and diversity, CRAFTS reaches the lowest PLIP-FID (11.32), improving over Stable Diffusion (15.82), Imagen (21.21) and StyleGAN-T (17.81) by ~28%, ~47% and ~36%, respectively (Fig. 2b). This reduction indicates that CRAFTS samples occupy a feature distribution substantially closer to real H&E images, reflecting fewer generative artefacts and richer morphological variability rather than mode-collapsed textures. Second, CRAFTS produces synthetic images that are semantically more aligned with real pathology: PLIP-I rises to 85.74%, exceeding competing models by 3.63–7.24 percentage points (Fig. 2c), consistent with improved recovery of diagnostically meaningful structures rather than superficial stain statistics. Crucially, the model also strengthens text-conditioned synthesis, attaining the highest PLIP-T (29.24%), with gains of 0.79 and 1.23 points over Stable Diffusion and Imagen, and a much larger 4.17-point margin over StyleGAN-T (Fig. 2d). The concurrent improvements in PLIP-I and PLIP-T suggest that CRAFTS not only generates visually faithful slides, but does so in a way that is tightly constrained by the prompt, yielding images whose nuclear atypia, architectural disorganisation and stromal context are more consistently instantiated from text. Together, these results position CRAFTS as superior on both axes required for clinical-grade synthesis: high-fidelity, diverse pathology, and reliably grounded text-image correspondence.

**Pathologist discrimination of real versus synthetic images.** We next conducted a blinded clinical realism study with three board-certified pathologists operating in consensus. A total of 500 real H&E tiles were randomly sampled from the held-out test cohorts, and 500 synthetic tiles were generated per method under identical prompt settings; for each comparison round, real and synthetic images were pooled, shuffled, and presented without source identifiers. Pathologists independently judged whether each image was real or synthetic, focusing on diagnostic plausibility at both architectural and cytologic scales, including nuclear sharpness, chromatin contrast, gland or nest organisation, stromal context, and staining coherence. Their decisions were aggregated to compute F1 scores for each method (Fig. 2e). CRAFTS yielded the lowest discriminability (F1 66.39%), compared with Stable Diffusion (68.92%), Imagen (71.87%), and StyleGAN-T (81.00%), indicating that CRAFTS images most closely approximate the perceptual and histomorphological statistics of real tissue. In practice, CRAFTS samples frequently achieved near-realistic nuclear detail and consistent microenvironmental structure, whereas baselines were more readily flagged due to blur, inconsistent hematoxylin-eosin balance, or non-biological textural priors, making them easier to separate from genuine slides.

**Pathologist ranking the semantic accuracy of synthetic images.** To evaluate conditioning faithfulness, we conducted a second blinded study focused on text-image correspondence. We randomly selected 500 pathology descriptions from the test set, each encoding explicit histological cues (for example, pleomorphism, keratin pearls, spindle-cell fascicles, desmoplasia, or infiltrative growth). For each description, every method generated one synthetic image, yielding 500 images per method. The three pathologists, again unaware of model identity, jointly reviewed each prompt-image group and assigned a semantic-alignment score on a 4-point ordinal scale (1–4 per image; 4 = best match, 1 = poorest match), reflecting how completely and accurately the generated morphology instantiated the textual semantics rather than overall aesthetics. This protocol penalised generic H&E-like outputs and rewarded faithful rendering of prompt-specified diagnostic motifs and their spatial organisation. CRAFTS achieved the highest mean alignment score (3.27), substantially exceeding Stable Diffusion (2.23), Imagen (2.46) and StyleGAN-T (2.05) (Fig. 2f). The margin reflects CRAFTS's higher consistency in reproducing key semantic targets, including epithelial architecture with stromal invasion, squamous differentiation with keratinisation, or organised spindle cell patterns, while competing models more often drifted toward nonspecific textures or missed prompt critical cellular features.

**Cancer-type feature distribution.** To further verify that CRAFTS preserves cancer-specific morphology under identical semantic inputs, we conducted a prompt-matched feature analysis followed by t-SNE visualisation. We randomly sampled a subset of cancer-annotated text descriptions from the test set and, for each description, generated one synthetic image with each method, ensuring strict one-to-one correspondence of prompts across models. Deep representations of these synthetics were extracted using the Prov-GigaPath[50] foundation model, and the resulting features were projected into two dimensions via t-SNE to reveal classwise structure (Fig. 2g). The t-SNE map shows that CRAFTS synthetics form compact, well-separated clusters aligned with their target cancer types, indicating faithful translation from text to type-defining histomorphology. In contrast, Stable Diffusion exhibits noticeably looser clusters with partial overlap, and Imagen and StyleGAN-T display substantial inter-class mixing and collapsed regions, consistent with semantic drift even when conditioned on the same prompts. This qualitative separation is supported quantitatively by silhouette coefficients: CRAFTS achieves a strong positive score (34.37%), markedly higher than Stable Diffusion (13.68%), while Imagen (−9.98%) and StyleGAN-T (−19.88%) yield negative separability, implying that many of their prompt-conditioned images embed closer to non-target classes than to their intended class. Together, the t-SNE topology and separability statistics demonstrate that CRAFTS maintains cancer-type discriminative structure in a pathology-aware feature space, reinforcing both its superior synthesis quality and its reliability for downstream augmentation.

**Downstream tasks using synthetic images**

To quantify the practical value of CRAFTS in clinically relevant learning pipelines, we systematically benchmark its synthetic images as a controllable data augmentation source across four downstream scenarios that collectively span classification, cross-modal retrieval, self-supervised learning and visual question answering (Fig. 3). For every task, we start from the standard real training splits and progressively enrich them with synthetic tiles generated from held out pathology descriptions. Synthetic images are injected at increasing synthetic-to-real ratios, enabling us to probe performance under mild augmentation as well as under aggressive settings that emulate rare cancer scarcity or severe class imbalance. To ensure a fair, prompt-conditioned comparison, the same set of textual inputs is used to drive generation for all methods, including Stable Diffusion, Imagen, and StyleGAN-T, and synthetic samples are matched in number, resolution, and preprocessing. Downstream models are trained with identical architectures, optimisation schedules, and evaluation protocols across augmentation sources, so that any performance difference can be attributed to the underlying quality and semantic faithfulness of the synthetic images.

**Classification.** We evaluated the impact of synthetic data augmentation on diagnostic classification using four publicly available pathology datasets: BACH[53], BRACS[54], BreakHis[55], and Lunghist[56], each representing distinct tissue types and grading systems (Fig. 3a). Multi-class classifiers based on the ViT[57] architecture were trained on real images and progressively augmented with synthetic images at varying ratios, ranging from 10% to 1,000%. The results consistently show that CRAFTS enhances model performance as the proportion of synthetic images increases, with accuracy improving across all datasets. For instance, in the BACH dataset, a 1:1 synthetic-to-real ratio yields an accuracy of 46.29%, which increases to 50.11% at a 10:1 ratio. Similarly, in BRACS, accuracy reaches 51.06% at a 1:1 ratio and continues to improve as the synthetic images ratio rises, demonstrating sustained performance gains. In BreakHis, CRAFTS achieves 60.33% accuracy at a 10:1 ratio, outperforming other methods. In Lunghist, CRAFTS attains 53.25% accuracy at a 10:1 synthetic-to-real ratio, showcasing the model's robustness even when a large proportion of synthetic images is used.

In contrast, models trained on synthetic images generated by other generative methods, such as Stable Diffusion, Imagen, and StyleGAN-T, do not exhibit the same sustained improvements. While CRAFTS shows a continuous increase in performance, other models either plateau or decline as the synthetic image ratio increases. This disparity can be attributed to CRAFTS's ability to generate synthetic images that are both visually realistic and semantically accurate, capturing essential diagnostic features like tissue structure, cellular morphology, and disease-specific patterns. In contrast, other generative models often produce visually plausible images but fail to preserve the semantic integrity required for pathology tasks, leading to a degradation in performance as more synthetic images are added. The consistent performance improvements observed with CRAFTS highlight its ability to generate high-quality, semantically grounded synthetic images, significantly enhancing the performance across various clinical pathology applications.

**Cross-modal retrieval.** We evaluated the impact of synthetic images on cross-modal retrieval tasks using the ARCH[59] and PathMMU[60] datasets, focusing on both text-to-image (T2I) and image-to-text (I2T) retrieval with the standard CLIP[58] model with the ViT architecture (Fig. 3b). In both datasets, CRAFTS-augmented models consistently outperformed others as the synthetic images ratio increased. For the ARCH dataset, retrieval performance (measured by R@5) improved as the synthetic images ratio increased. In the T2I task, performance rose from 32.65% at a 1:1 synthetic-to-real ratio to 33.25%, and further increased to 35.92% at a 10:1 ratio. In the I2T task, performance improved from 30.22% to 32.04% at a 1:1 ratio and reached 35.44% at a 10:1 ratio. Similarly, on the PathMMU dataset, the T2I task increased from 6.56% to 8.23%, while the I2T task improved from 6.56% to 8.10% at a 10:1 ratio. In contrast, models trained on synthetic images generated by other generative methods, such as Stable Diffusion, Imagen, and StyleGAN-T, either stagnated or showed a noticeable decline in performance as more synthetic images were added. This disparity highlights the superior semantic alignment of CRAFTS-generated images for cross-modal retrieval tasks. CRAFTS consistently produces synthetic images that accurately capture the semantic features required for effective retrieval, such as tissue structures, cellular morphology, and disease-specific patterns. These features are crucial for maintaining meaningful relationships between textual descriptions and image content, ensuring that retrieval performance improves as more synthetic images are incorporated. In contrast, other generative models like Stable Diffusion, Imagen, and StyleGAN-T often produce images that, while visually plausible, lack the precise semantic fidelity needed for reliable retrieval. This semantic misalignment leads to stagnation or even degradation in performance as the synthetic images ratio increases.

**Self-supervised learning.** We evaluated the impact of synthetic images on self-supervised learning tasks using the BRACS and ColonPath[66] datasets, with downstream tasks testing on BreakHis, Pcam[61], and BACH for BRACS, and MHIST[62], HMNIST[63], and NCTCRC[64] for ColonPath. The self-supervised model, SimSiam[65], was employed to assess the effect of synthetic data on generalization. As shown in Fig. 3c, incorporating synthetic images from CRAFTS led to significant performance improvements. On the BRACS dataset, adding synthetic data at a 1:1 ratio increased classification accuracy on the downstream BreakHis task from 35.88% to 38.17%, and further increased to 47.07% at a 10:1 ratio. Similarly, on the ColonPath dataset, accuracy on the downstream MHIST classification task improved from 59.47% to 59.88% at a 1:1 ratio, and to 66.33% at a 10:1 ratio. A similar trend was observed for the downstream tasks on Pcam and BACH for BRACS, HMNIST and NCTCRC for ColonPath, where synthetic data from CRAFTS consistently improved performance. These results demonstrate that as synthetic data from CRAFTS increases, the performance improves, with the synthetic images effectively augmenting the learning process and boosting downstream task accuracy. In contrast, when synthetic data from other generative methods, such as Stable Diffusion, Imagen, and StyleGAN-T, was used, the performance improvements were either minimal or stagnated, and in some cases, the performance even decreased as the synthetic data ratio increased. This discrepancy underscores the superior semantic alignment of CRAFTS-generated images. CRAFTS consistently produces synthetic images that accurately reflect key pathological features, such as tissue structures, cellular morphology, and disease-specific patterns, which are critical for self-supervised learning tasks. Other methods, however, often generate images that, while visually plausible, lack the semantic precision necessary for effective feature learning, leading to diminished improvements or performance declines as more synthetic data is added.

**Visual question answering.** In Visual Question Answering (VQA) tasks using the PatchVQA[67] and ARCH datasets, we assessed the impact of synthetic images generated by CRAFTS. The VQA model employed T5[68] as the text encoder and ViT as the image encoder, projecting visual features into the text space, which were then integrated with text embeddings to generate answers. As shown in Fig. 3d, incorporating CRAFTS-generated synthetic images led to significant performance improvements. For PatchVQA, the METEOR score increased from 9.50% to 10.77% at a 1:1 synthetic-to-real ratio, and further to 16.60% at a 10:1 ratio, surpassing models trained with synthetic data from other generative methods. Similarly, on the ARCH dataset, the METEOR score for CRAFTS-augmented models improved from 35.62% to 37.04% at a 1:1 synthetic-to-real ratio, and further to 38.97% at a 10:1 ratio, outperforming models trained with synthetic data from alternative methods. These results underscore CRAFTS's ability to generate semantically accurate images that capture key diagnostic features, such as tissue structures and cellular morphology, which are critical for enhancing VQA task performance.

**Structure-controllable image generation based on CRAFTS**

To evaluate whether the semantic priors encoded in CRAFTS can be exploited for controllable pathology image synthesis, we coupled the pre-trained CRAFTS with ControlNet and compared it with a ControlNet-conditioned Stable Diffusion model under identical settings (Fig. 4). For structure-aware control, nuclear segmentation masks from the GLySAC[69] and CoNSeP[70] datasets were used as spatial conditions, and both models were prompted to "translate the given nuclear segmentation mask into a realistic H&E-stained histology image, generating hematoxylin-stained nuclei and eosin-stained cytoplasm/background with accurate tissue texture and color distribution". For phenotype-aware control, immunofluorescence images from the HEMIT[71] and SHIFT[72] datasets were used as conditions with an analogous prompt instructing translation from fluorescence to realistic H&E images. The conditional image, real H&E counterpart, and synthetic outputs were compared quantitatively using a pathology-aware PLIP-FID score, as well as Structure Similarity Index Measure (SSIM), Mean-Squared Error (MSE), Normalized Cross-Correlation (NCC), and Peak Signal-to-Noise Ratio (PSNR) against the corresponding real slide patches, thereby jointly probing semantic consistency, structural fidelity, and low-level image quality.

**Mask to H&E image translation.** When conditioned on nuclear segmentation masks, CRAFTS consistently produced images that were both closer to real H&E patches and more faithful to the conditioning geometry than Stable Diffusion (Fig. 4a,b). On GLySAC, CRAFTS achieved a lower PLIP-FID than Stable Diffusion (16.73 vs 17.13) and showed improved SSIM (18.18% vs 16.93%), reduced MSE (7.13 vs 8.47; ~16% reduction), and higher NCC and PSNR (46.49% vs 40.25% and 11.58 vs 10.82, respectively). Compared with Stable Diffusion, CRAFTS showed amplified advantages on the more challenging CoNSeP dataset, with PLIP-FID reduced from 21.78 to 16.82, SSIM increased from 30.15% to 33.75%, MSE reduced from 3.55% to 2.49%, NCC increased from 64.00% to 67.92%, and PSNR from 15.32 to 17.38. Qualitatively, in the GLySAC examples (top row of Fig. 4a), CRAFTS respects the segmentation mask by placing nuclei at the correct locations and preserving their anisotropic, elongated morphology and nuclear-cytoplasmic boundaries, while Stable Diffusion frequently hallucinates nuclei outside the masked regions, collapses closely apposed nuclei into amorphous basophilic clumps, and distorts glandular lumina. In the CoNSeP examples (second row), CRAFTS better reconstructs the colonic gland architecture, including continuous apical borders, sharp luminal contours, and realistic nuclear stratification along the gland wall; in contrast, Stable Diffusion introduces fragmented epithelium, smeared chromatin texture, and locally inconsistent nuclear sizes. In the zoomed-in patches, CRAFTS produces crisp chromatin with visible nucleoli, smooth eosinophilic stromal background, and balanced hematoxylin-eosin contrast, whereas Stable Diffusion exhibits blurred nuclear profiles, oversaturated eosin staining, and patchy textural artifacts. Together, these quantitative and qualitative observations indicate that CRAFTS not only maintains higher image quality but also preserves a closer correspondence between the conditioned nuclear layout, the generated histology, and the textual prompt.

**Fluorescence to H&E image translation.** A similar pattern is observed when controlling generation with fluorescence images (Fig. 4c,d). On HEMIT, CRAFTS attains lower PLIP-FID than Stable Diffusion (16.78 vs 18.79), higher SSIM (37.55% vs 33.03%), substantially reduced MSE (3.09 vs 4.74), and improved NCC (64.86% vs 58.79%) and PSNR (15.33 vs 13.57). On SHIFT, CRAFTS again outperforms Stable Diffusion across all metrics, with PLIP-FID of 21.11 vs 25.30, SSIM of 40.50% vs 39.51%, MSE of 2.00% vs 2.38%, NCC of 62.29% vs 59.41% and PSNR of 17.31 vs 16.45. In the HEMIT examples (top row of Fig. 4c), CRAFTS accurately maps high-intensity fluorescent clusters to densely packed hyperchromatic nuclei and preserves the radial arrangement of epithelial cells around glandular lumina, while Stable Diffusion often breaks this correspondence, yielding misplaced or missing nuclear groups and irregular luminal borders. In the SHIFT examples (second row of Fig. 4c), CRAFTS generates finely granular chromatin, well-separated nuclei with appropriate size dispersion, and coherent stromal and epithelial textures, whereas Stable Diffusion produces softer, slightly washed-out nuclei, discontinuous epithelial layers, and less convincing stromal collagen patterns. The zoomed-in regions highlight that CRAFTS more faithfully converts subtle intensity gradients in the fluorescence images into variations in nuclear density and staining intensity in H&E space, resulting in synthetic images that are simultaneously of higher perceptual quality and more tightly aligned with the fluorescence-encoded biology and the accompanying textual instructions.

## Discussion

CRAFTS establishes a pathology-focused generative foundation model that advances text-to-image synthesis in computational pathology. By coupling large-scale pre-training on heterogeneous, low-quality medical corpora with cancer-type-informed fine-tuning on high-quality image-caption pairs, the framework addresses a central obstacle in the field, namely the chronic scarcity and imbalance of expert-annotated pathological images. The correlation-regulated alignment mechanism is specifically designed to suppress semantic and morphological hallucinations while preserving diagnostic consistency between text and image. This design enables CRAFTS to generate morphologically diverse yet clinically coherent images across 30 cancer types. A comprehensive evaluation using quantitative fidelity metrics and expert pathologist review confirms that the synthetic images capture essential diagnostic features, underscoring the practical relevance of generative models for pathology.

In addition to high-fidelity synthesis, CRAFTS provides substantial utility for downstream clinical applications. Synthetic data generated by CRAFTS improves performance across multiple tasks, including cancer classification, cross-modal retrieval, self-supervised learning, and visual question answering. These improvements stem from the model's ability to reproduce fine-grained morphological variations while maintaining semantic consistency with textual descriptions. The controllability of CRAFTS, achieved through few-shot conditioning and its integration with ControlNet, further broadens its applicability. Conditioning on nucleus-level masks or fluorescence images enables fine manipulation of tissue architecture and cellular context, making the model particularly effective for augmenting datasets containing rare cancers or underrepresented histological patterns.

Despite these advances, several challenges remain. At present, CRAFTS generates image patches at 512×512 resolution due to memory constraints. Given that clinical interpretation often depends on contextual information present in gigapixel whole-slide images, future research should explore hierarchical generation strategies or memory-efficient architectures that can scale synthesis to whole-slide resolution while preserving cellular detail. Another challenge lies in aligning complex pathological descriptions with their visual counterparts. As the richness and length of pathology reports increase, current text encoders may struggle to represent subtle diagnostic cues or structured clinical information. Refining multimodal alignment, perhaps through domain-specific language models or structured clinical knowledge, will be essential for producing images that more accurately reflect diagnostic intent and biological context.

The broader implications of CRAFTS extend beyond generative modeling itself. By offering a scalable method for producing diverse, pathology-grade synthetic images, CRAFTS addresses longstanding limitations arising from limited access to high-quality and balanced datasets. Its capacity to enrich training distributions, particularly for rare cancers, can improve diagnostic robustness, reduce biases caused by data imbalance, and support the development of clinically reliable AI systems. The use of text-conditioned synthesis also provides a pathway toward generating images that encode both morphological detail and disease-specific semantic information, thereby enhancing the clinical relevance of synthetic datasets. As generative models become more deeply integrated into clinical research workflows, frameworks such as CRAFTS may help standardize data generation, accelerate algorithm development, and promote more equitable diagnostic innovation.

Overall, CRAFTS represents an important step toward clinically grounded generative foundation models for oncology. Its integration of semantic conditioning with high-fidelity image synthesis opens new opportunities for dataset augmentation, diagnostic model development, and controlled exploration of disease morphology. Continued progress in scaling, multimodal alignment, and clinical integration will further establish generative modeling as a core component of next-generation computational pathology.

## Methods

### Training datasets

We trained CRAFTS on a two-stage corpus that combines broad, heterogeneous visual-semantic knowledge with high-fidelity, cancer-focused supervision. The pre-training stage uses approximately 1.2 million image-caption pairs assembled from four publicly available medical vision-language resources: 13.2k pairs from PubMedVision[41], 42k from PMC-OA[43], 17k from PMC-VQA[42], and 1.0 million from Quilt-1M[44] (Fig. 1a). These data cover a wide spectrum of organs, imaging modalities, and descriptive styles, and provide rich but noisy semantic supervision without explicit cancer type labels. For fine tuning, we employ a high quality, TCGA based corpus of 1.6 million pathology image-text pairs derived from the PathGen-1.6M dataset[45] (Fig. 1b). Each image patch is extracted from diagnostic whole slide images and paired with a concise, morphology oriented caption and a curated label from one of 30 cancer types, yielding a focused, pathology specific distribution that anchors CRAFTS to real clinical phenotypes.

**Pre-training dataset.** The pre-training dataset is designed to maximise semantic coverage rather than visual fidelity. From PubMedVision, we sample 13.2k medically relevant image-caption pairs obtained by filtering PubMed figures and using a multimodal large language model to generate detailed, image-grounded captions and question-answer pairs across diverse body parts and modalities. From PMC-OA, we draw 42k biomedical image-caption pairs constructed via subfigure detection and subcaption alignment across more than two million PubMed Central articles, which substantially broaden the range of diagnostic procedures and disease entities, including radiology, microscopy, and generic biomedical illustrations. We further include 17k pathology leaning pairs adapted from PMC-VQA, where we repurpose VQA triples into descriptive captions that emphasise anatomical location, imaging modality, and local pathology. Finally, Quilt-1M provides 1.0 million histopathology image-caption pairs mined primarily from educational YouTube videos, social media, and open-access articles, with captions that describe regions of interest, subpathology, and magnification across multiple organ systems. Although these sources collectively encode broad medical and histopathological knowledge, the images are often compressed, stylistically heterogeneous, and only loosely aligned with their accompanying text, and they lack consistent cancer type annotations. In CRAFTS, we therefore use this corpus to establish robust text-image grounding and to expose the model to diverse morphological and semantic patterns rather than to define the final generative distribution.

**Fine-tuning dataset.** The fine-tuning dataset provides complementary properties: high image quality, explicit cancer labels, and focused pathological content. PathGen-1.6M is constructed from approximately 7,300 TCGA whole slide images, where a multi-agent pipeline first retrieves representative patches using a pathology-specific CLIP model and then employs a large multimodal pathology model to generate detailed descriptions that are subsequently revised and summarised. The resulting 1.6 million image-caption pairs capture cellular morphology, tissue architecture, and key diagnostic cues at the patch level while remaining tightly coupled to refined slide-level reports. For CRAFTS, we map each patch to one of 30 consolidated cancer types, yielding a high-resolution, category-informed supervision signal that constrains the model to clinically realistic feature manifolds and provides explicit priors for cancer-aware conditioning.

In contrast to the broad but noisy pre-training corpus, this dataset is visually clean, pathology-only, and label-rich, which is used to specialise CRAFTS for oncologic tissue synthesis.

## CRAFTS's framework and implementation details

CRAFTS is built on a pathology-specialised latent diffusion architecture designed to generate clinically faithful, semantically grounded histopathological images (Fig. 1c). The framework introduces a two-stage training paradigm that progressively couples broad semantic grounding with high-fidelity cancer-type supervision. In the pre-training stage, the model acquires coarse yet scalable cross-modal correspondences from a dataset of 1.2 million weakly paired medical image-text pairs. This exposure enables the diffusion backbone to internalise a wide spectrum of pathological descriptors, imaging styles, and narrative structures. Fine-tuning on 1.6 million TCGA-derived pairs with explicit cancer-type annotations then injects diagnostic priors and constrains the generative process to well-defined histomorphological manifolds.

CRAFTS is a correlation-regulated alignment framework that operates across both stages to stabilise text-image grounding and mitigate semantic or morphological hallucinations. A semantic consistency constraint aligns relational similarity structures between generated image latents and their conditioning texts, ensuring faithful translation of descriptive cues into visual patterns. During fine-tuning, a complementary category-guided constraint further aligns image latents with cancer-type embeddings, thereby embedding explicit diagnostic priors into the generative geometry. Together, these mechanisms modulate the denoising trajectory of the diffusion model, enabling CRAFTS to synthesise pathology images that are not only visually plausible but also diagnostically interpretable and precisely aligned with textual instructions.

**Latent diffusion backbone for pathological images synthesis.** CRAFTS is instantiated as a latent diffusion model operating in the compressed representation space of a high-capacity variational autoencoder (VAE). Given an input H&E patch $x_0 \in \mathbb{R}^{H \times W \times 3}$, where $H = W = 512$, the VAE encoder $E_{VAE}$ maps it to a latent feature map $z_0 = E_{VAE}(x_0)$. Working in this latent space allows the diffusion backbone to model high-level tissue structure and cellular morphology with substantially reduced computational cost, while the VAE decoder $D_{VAE}$ later reconstructs a full-resolution synthetic patch $\hat{x}_0 = D_{VAE}(z_0)$.

To train the generative backbone, we corrupt $z_0$ with a Markovian forward diffusion process that gradually adds Gaussian noise over $T$ timesteps. At each timestep $t \in \{1, \ldots, T\}$, the noised latent $z_t$ is obtained from $z_{t-1}$ as

$$q(z_t | z_{t-1}) = N(z_t; \sqrt{1 - \beta_t} z_{t-1}, \beta_t \mathbf{I}), \quad (1)$$

where $\beta_t \in (0,1)$ is the variance of the injected noise at step $t$, and $\mathbf{I}$ is the identity matrix in latent space. We use a linear noise schedule

$$\beta_t = \beta_{min} + \frac{t}{T}(\beta_{max} - \beta_{min}), \quad (2)$$

with $\beta_{min}$ and $\beta_{max}$ denoting the minimum and maximum noise levels, respectively. Defining $\alpha_t = 1 - \beta_t$ and the cumulative product $\bar{\alpha}_t = \prod_{s=1}^{t} \alpha_s$, the closed-form perturbation from the clean latent $z_0$ to any timestep $t$ can be written as

$$q(z_t | z_0) = N(z_t; \sqrt{\bar{\alpha}_t} z_0, (1 - \bar{\alpha}_t) \mathbf{I}), \quad (3)$$

so that a sample at timestep $t$ is generated as

$$z_t = \sqrt{\bar{\alpha}_t} z_0 + (1 - \bar{\alpha}_t) \varepsilon, \quad (4)$$

with $\varepsilon \sim N(0, \mathbf{I})$. Intuitively, early timesteps preserve glandular architecture and nuclear topology, while late timesteps approach pure noise and discard most histological structure.

The reverse process is learned by a U-Net denoiser $\varepsilon_\theta$ that predicts the corruption noise from a noised latent $z_t$, the diffusion timestep $t$, and textual conditioning $c$. The conditioning vector $c$ is a sequence of token embeddings from the text encoder that captures the diagnostic descriptive cues. During training, we minimise the standard denoising diffusion objective

$$\mathcal{L}_{diff} = \mathbb{E}_{z_0,\varepsilon,t,c}[\|\varepsilon - \varepsilon_\theta(z_t, t, c)\|_2^2], \quad (5)$$

encouraging $\varepsilon_\theta$ to accurately recover the injected noise $\varepsilon$ and thereby learn the score function of the latent distribution conditioned on the pathology description.

At inference time, image synthesis starts from pure Gaussian noise $z_T \sim N(0, I)$ and iteratively applies a learned reverse-diffusion update for $t = T, \dots, 1$:

$$z_{t-1} = \frac{1}{\sqrt{\alpha_t}}\left(z_t - \frac{\beta_t}{\sqrt{1-\bar{\alpha}_t}}\varepsilon_\theta(z_t, t, c)\right) + \sigma_t z, \quad (6)$$

where $z \sim N(0, I)$ and $\sigma_t^2$ controls the residual stochasticity of the reverse step. After $T$ denoising steps, the final latent $z_0$ is decoded by $D_{VAE}$ into a synthetic H&E patch whose nuclear detail, stromal context, and architectural organisation are jointly determined by the conditioning text.

To tightly couple the generative process with pathological semantics, cross-attention layers are inserted at multiple resolutions of the U-Net. For a given U-Net feature map $F_t^l$ at layer $l$, we construct query vectors $Q^l$ via a linear projection of $F_t^l$, while keys $K$ and values $V$ are obtained from the token sequence from the text encoder. Each cross-attention block computes

$$\text{CrossAttn}(Q^l, K, V) = \text{softmax}\left(\frac{Q^l K^T}{\sqrt{d_k}}\right)V, \quad (7)$$

where $d_k$ is the key dimension. Low-resolution layers mainly integrate global cues describing tissue-level architecture, whereas high-resolution layers refine cell-level details such as nuclear atypia, mitotic figures, and chromatin texture. This multi-scale conditioning design ensures that the latent diffusion backbone does not merely reproduce H&E-like textures, but instead generates images whose micro- and macro-morphology follow the clinically meaningful semantics encoded in the pathological descriptions and cancer-type priors.

**Semantic consistency for mitigating hallucinations.** In the field of pathological image synthesis, the reliability of generative models depends on their ability to strictly adhere to textual semantics. Standard generative models align individual images with their associated textual descriptions to achieve text-to-image synthesis. However, textual descriptions are typically highly summary and struggle to impose complete, comprehensive semantic constraints on complex tissue morphologies. This semantic under-specification leads to significant uncertainty in the generation process, resulting in a mapping from abstract text to concrete images that involves numerous unconstrained stochastic inferences. Furthermore, within these undefined semantic blind spots, models are highly susceptible to interference from non-standardized terminology or noise, leading to the fabrication of biologically baseless textures and, consequently, severe hallucinations. To address this limitation, this study establishes a semantic consistency constraint mechanism based on intra-batch sample correlations. This mechanism establishes a synchronized evolutionary relationship between visual and linguistic feature spaces, ensuring that semantically highly correlated descriptions generate images exhibiting commensurate similarities in the visual feature space, while texts with significant semantic differences maintain corresponding distances in visual representation. This approach no longer relies solely on the absolute control of a single text over an image but instead introduces a relational structure within a batch as additional supervisory information, thereby leveraging the inherent stability of pathological knowledge to rectify stochastic deviations in the generation process.

Specifically, for a training batch containing $B$ samples, we define $T \in \mathbb{R}^{B \times d_t}$ as the set of semantic features extracted by the text encoder, and $Z \in \mathbb{R}^{B \times d_v}$ as the set of visual latent variables in the image generation process, where $d_t$ and $d_v$ denote the feature dimensions, respectively. To quantify the semantic association between samples, we first compute the pairwise cosine similarity matrix for text features, denoted as $M^T \in$ $\mathbb{R}^{B \times B}$, where the element $M^T_{ij}$ at the $i$-th row and $j$-th column characterizes the strength of the semantic association between the pathological description of the $i$-th sample and that of the $j$-th sample. The calculation is formulated as follows:

$$M^T_{ij} = \frac{t_i \cdot t_j}{\|t_i\| \|t_j\|}. \tag{8}$$

Similarly, we compute the visual similarity matrix $M^Z \in \mathbb{R}^{B \times B}$ among the latent variables of the generated images to characterize the relative distribution relationships of the synthetic images in the feature space:

$$M^Z_{ij} = \frac{z_i \cdot z_j}{\|z_i\| \|z_j\|}. \tag{9}$$

Here, $t_i$ and $z_i$ represent the text feature vector and the image feature of the $i$-th sample, respectively. Based on these definitions, the semantic consistency loss $\mathcal{L}_{corr}$ is defined as the divergence between the two similarity matrices, constraining the structural evolution of the visual space by minimizing this difference:

$$\mathcal{L}_{corr} = \frac{1}{B^2} \sum_{i=1}^{B} \sum_{j=1}^{B} \|M^Z_{ij} - M^T_{ij}\|^2. \tag{10}$$

This relationship-based constraint mechanism essentially acts as a structural filter. Since non-standardized terminology or random noise typically lacks stable cross-sample correlations, they fail to form a consistent similarity structure at the batch level. Therefore, by minimizing the relational difference between the two feature spaces, the model effectively ignores stochastic noise that cannot form stable structures and instead focuses on robust features that are biologically consistent. This ensures that the generated pathological images are not only realistic in textural detail but also faithful to the classification and description systems of pathology in their deep semantic logic.

**Category guidance for injecting diagnostic priors.** Although semantic consistency constraints effectively regulate the mapping process from text to image and suppress generative hallucinations induced by semantic uncertainty and random noise, precise pathological image generation still requires distinct biological categorical attribution to ensure consistency of synthetic samples at the clinical diagnostic level. Textual descriptions often focus on microscopic cytological features or tissue structures, yet lack clear definitions of macroscopic disease categories. This semantic ambiguity between microscopic descriptions and macroscopic diagnoses easily leads the model to confuse different cancer subtypes with similar morphological features during generation. For instance, certain lung and colorectal adenocarcinomas are extremely similar in their glandular structure, making them difficult to distinguish solely on morphological grounds. To bridge this gap, this study introduces a semantic-aware category-guidance strategy during fine-tuning. This strategy aims to inject distinct cancer-category information into the generation process, leveraging global diagnostic priors inherent in category labels to guide visual latent variables toward specific cancer-category manifolds, thereby ensuring that synthetic images possess accurate biological characterization alongside rich morphological details. Furthermore, rather than rigidly forcing all images toward their respective category centers, this strategy constructs a differentiated elastic constraint mechanism by assessing the degree of support for a specific diagnostic category in the text description, thereby adaptively regulating the guidance intensity of category priors in the generative model.

Specifically, the model first evaluates the semantic association strength between the text description and a specific cancer category to determine whether the sample exhibits typical diagnostic features, and subsequently uses this association strength as a confidence weight to adaptively adjust the guidance from category information during the generation process. For sample $k$, we first compute the cosine similarity $s_k$ between its text description feature $t_k$ and the corresponding cancer category feature $c_k$ to quantify the degree of support of the text for that diagnostic category:

$$s_k = \cos(c_k, t_k). \tag{11}$$

Subsequently, to construct robust guidance at the batch level, we define the category typicality score for sample pair $(i,j)$ as $A_{ij} = s_i + s_j$. This score intuitively reflects the clarity of the sample pair regarding category attribution, where a higher score implies that both samples possess typical category features. To eliminate data fluctuations between batches and achieve adaptive regulation, we standardize this score using the global mean $\mu$ and standard deviation $\sigma$ of the entire fine-tuning dataset and transform it into a dynamic weight $W_{ij}$ through the Sigmoid function:

$$W_{ij} = \frac{1}{1 + \exp\left(-\frac{A_{ij} - \mu}{\sigma}\right)}. \tag{12}$$

Based on this weight, the category guidance loss $\mathcal{L}_{cate}$ is defined as the weighted difference between the image similarity matrix $M^Z \in \mathbb{R}^{B \times B}$ and the category similarity matrix $M^C \in \mathbb{R}^{B \times B}$:

$$\mathcal{L}_{cate} = \frac{1}{B^2} \sum_{i=1}^{B} \sum_{j=1}^{B} W_{ij} \left\| M_{ij}^Z - M_{ij}^C \right\|^2. \tag{13}$$

When the text description is highly specific and clearly points to a specific cancer type, the model imposes stronger constraints, forcing the generated images to strictly adhere to the feature distribution of that category. Conversely, when the description is vague or possesses atypical features, the model moderately reduces the intensity of category constraints, allowing the generated images to retain greater morphological diversity. In this manner, the category guidance mechanism not only effectively sharpens the feature boundaries between different cancer subtypes and prevents cross-category feature confusion but also avoids the loss of subtle heterogeneous features crucial for clinical diagnosis due to over-regularization.

Together, the latent diffusion backbone and the correlation-regulated alignment framework endow CRAFTS with a dual-stage grounding strategy that integrates broad semantic understanding with precise diagnostic priors, robust hallucination suppression via relational consistency constraints, and category-aware control, enabling faithful synthesis of cancer-specific phenotypes. These components collectively advance the model from producing visually plausible images to generating clinically grounded, diagnostically precise, and semantically coherent pathology samples, establishing CRAFTS as a foundation model capable of accurate and controllable pathological image synthesis.

**Training details for CRAFTS.** CRAFTS was fine-tuned using pre-trained U-Net weights from the general domain, specifically the Stable Diffusion (SD) version 1.5 architecture and CLIP (ViT-large-patch14) text encoder obtained from HuggingFace. Leveraging this backbone, CRAFTS comprises approximately 1.0 billion trainable parameters across the denoising U-Net and text encoder. The model was fine-tuned on a high-performance cluster with ten NVIDIA RTX A6000 GPUs. Images were resized to 512×512 pixels and processed in batches of 320. The training consisted of two phases: the first with 1.2 million steps (approximately 2 days) and the second with 1.6 million steps (approximately 4 days), completing in 6 days across 5 epochs. During inference, we employed a classifier-free guidance scale of 7.5 and set the number of diffusion steps to 50, using the default noise scheduler. The *transformers* and *diffusers* libraries were utilized for model implementation, and the built-in safety checker was disabled to prevent false positives during medical image generation. The model was optimized for high-quality pathological image synthesis, balancing computational efficiency with accurate image generation.

## Metrics

**Accuracy (ACC).** Accuracy is a straightforward metric representing the proportion of correct predictions out of the total predictions. For multi-class classification, the formula for accuracy is:

$$\text{Accuracy} = \frac{\text{Number of Correct Predictions}}{\text{Total Number of Predictions}}. \tag{14}$$

Accuracy measures the overall correctness of a model's predictions. A higher ACC value indicates higher model accuracy.

**F1-Score.** The F1-score is a balanced metric that combines precision and recall into a single value, providing a more comprehensive evaluation of a model's performance, especially when dealing with imbalanced datasets. The F1-score is defined as:

$$\text{F1-score} = 2 \times \frac{Precision \times \text{Recall}}{Precision + \text{Recall}}, \tag{15}$$

where $\text{Precision} = \frac{\text{True Positives}}{\text{True Positives}+\text{False Positives}}$, $\text{Recall} = \frac{\text{True Positives}}{\text{True Positives}+\text{False Negatives}}$. The F1-score balances the trade-off between precision and recall; a higher F1-score indicates better model performance, particularly when both false positives and false negatives are critical.

**PLIP-FID.** To evaluate visual realism under pathology-specific semantics, we compute the Fréchet Inception Distance by replacing Inception features with PLIP embeddings. PLIP-FID measures the Fréchet distance between the multivariate Gaussian distributions of real and synthetic image features:

$$\text{PLIP-FID} = \|\mu_r - \mu_s\|_2^2 + \text{Tr}(\Sigma_r + \Sigma_s - 2(\Sigma_r\Sigma_s)^{1/2}), \tag{16}$$

where $\mu_r, \Sigma_r$ and $\mu_s, \Sigma_s$ denote feature means and covariances of real and generated images. Lower scores indicate higher pathology-specific realism.

**PLIP-I (Image-Image Similarity).** PLIP-I quantifies the image-level semantic consistency by computing the cosine similarity between PLIP-derived embeddings of real and synthetic patches:

$$\text{PLIP-I} = \frac{\langle f_r, f_s \rangle}{\|f_r\|\|f_s\|}, \tag{17}$$

where $f_r$ and $f_s$ denote PLIP image embeddings. Higher values indicate closer alignment in pathology-relevant visual semantics.

**PLIP-T (Text-Image Alignment).** PLIP-T measures cross-modal agreement between synthetic images and their conditioning textual descriptions. It computes cosine similarity between PLIP text embeddings $t$ and image embeddings $f_s$:

$$\text{PLIP-T} = \frac{\langle t, f_s \rangle}{\|t\|\|f_s\|}. \tag{18}$$

Higher PLIP-T indicates stronger text-image correspondence.

**Silhouette Coefficient.** To assess the cluster structure of generated phenotypes, we compute the Silhouette Coefficient:

$$s_i = \frac{b_i - a_i}{\max(a_i, b_i)}, \tag{19}$$

where $a_i$ s the mean intra-class distance for sample $i$, and $b_i$ s the smallest mean distance to other classes. The metric ranges from −1 to 1, with higher values indicating well-separated cancer-type clusters.

**Recall@5 (R@5).** For cross-modal retrieval, R@5 quantifies the proportion of queries (text or image) for which the correct counterpart appears in the top-5 retrieved results:

$$R@5 = \frac{1}{N} \sum_{i=1}^{N} \mathbb{I}(rank(i) \leq 5). \tag{20}$$

Higher R@5 reflects improved alignment between image and text representations.

**METEOR.** For visual question answering, we use METEOR to evaluate linguistic output fidelity. METEOR incorporates unigram recall, precision and paraphrase matching:

$$\text{METROR} = F_\alpha (1 - p), \tag{21}$$

where $F_\alpha$ is a harmonic mean of precision and recall and $p$ is a fragmentation penalty. Higher METEOR scores indicate more accurate and semantically coherent answers.

**Structural Similarity Index Measure (SSIM).** SSIM evaluates perceptual structural fidelity between synthetic and reference images:

$$\text{SSIM}(x, y) = \frac{(2\mu_x\mu_y + C_1)(2\sigma_{xy} + C_2)}{(\mu_x^2 + \mu_y^2 + C_1)(\sigma_x^2 + \sigma_y^2 + C_2)}, \tag{22}$$

where $\mu$, $\sigma^2$ and $\sigma_{xy}$ denote means, variances and cross-covariance. Higher SSIM indicates improved structural preservation.

**Mean-Squared Error (MSE).** MSE quantifies pixel-wise reconstruction error:

$$\text{MSE} = \frac{1}{n}\sum_{i=1}^{n}(x_i - y_i)^2. \tag{23}$$

Lower MSE reflects higher numerical accuracy.

**Normalized Cross-Correlation (NCC).** NCC assesses global morphological consistency between two images:

$$\text{NCC} = \frac{\sum_i (x_i - \mu_x)(y_i - \mu_y))}{\sqrt{\sum_i (x_i - \mu_x)^2}\sqrt{\sum_i (y_i - \mu_y)^2}}. \tag{24}$$

Values closer to 1 indicate stronger structural agreement.

**Peak Signal-to-Noise Ratio (PSNR).** PSNR measures reconstruction fidelity relative to pixel intensity range:

$$\text{PSNE} = 10\log_{10}\left(\frac{\text{MAX}^2}{\text{MSE}}\right), \tag{25}$$

where MAX denotes the maximum possible pixel value. Higher PSNR indicates better signal preservation and lower distortion.

## Compared methods

**Stable Diffusion.** We followed the guidelines and used the recommended default parameter setting on the diffusers GitHub repository: https://github.com/huggingface/diffusers.

**Imagen.** We followed the guidelines and used the recommended default parameter setting on the Imagen GitHub repository: https://github.com/lucidrains/imagen-pytorch.

**StyleGAN-T.** We used the code of StyleGAN-T from https://github.com/autonomousvision/stylegan-t. StyleGAN-T was run with the recommended default parameter.

To ensure strict experimental fairness, all comparison methods followed the same training process. The pre-training phase used an identical 1.2 million-image dataset, while the fine-tuning phase was conducted on a consistent 1.6 million-image dataset, ensuring that performance differences were attributable solely to the model architecture rather than variations in the training procedure.

## Evaluation datasets

**BACH** consists of 400 H&E-stained microscopy images (2048 × 1536 pixels) and 30 FFPE WSIs of breast histology from the ICIAR 2018 challenge, annotated into four classes: normal, benign, in situ carcinoma, and invasive carcinoma. Microscopy images are labeled at the image level, while WSIs include pixel-level annotations for localization tasks.

**BRACS** consists of 547 H&E-stained FFPE WSIs, 4,561 regions of interest, and 13,907 cellular-level annotated nuclei from breast carcinoma cases at the University of Naples Federico II. Images are classified into seven subtypes: normal, non-malignant benign, atypical, ductal carcinoma in situ, invasive ductal carcinoma, invasive lobular carcinoma, and other malignant.

**BreakHis** consists of 7,909 H&E-stained microscopy images (700 × 460 pixels) of breast tumor tissue from 82 patients at the P&D Laboratory, Brazil, acquired at four magnification levels (40×, 100×, 200×, 400×). Cases are classified as benign (2,480 images; subtypes: adenosis, fibroadenoma, phyllodes tumor, tubular adenoma) or malignant (5,429 images; subtypes: ductal carcinoma, lobular carcinoma, mucinous carcinoma, papillary carcinoma).

**LungHist** consists of 691 high-resolution (1,200 × 1,600 pixels) H&E-stained histopathological lung images from 45 patients at Hospital Clínic de Barcelona. Images are categorized as adenocarcinoma (232), squamous cell carcinoma (233), or normal tissue (226), supporting deep learning tasks in pulmonary pathology.

**ARCH** consists of 7,772 image-caption pairs extracted from histopathology textbooks and journal articles, curated for multiple instance captioning. Each image is paired with descriptive captions providing dense supervision for computational pathology tasks such as representation learning.

**PathMMU** consists of 33,428 multimodal multiple-choice questions paired with 24,067 images from diverse pathology sources, validated by experts for understanding and reasoning benchmarks. Questions cover subfields including diagnostic interpretation, prognostic prediction, and therapeutic response, with explanations for correct answers.

**ColonPath** consists of 15,000 H&E-stained images from 4,995 real-world cases in the MedFMC benchmark, focused on colon tumor classification (benign vs. malignant). It supports foundation model adaptation for medical image classification tasks.

**PCam** consists of 327,680 color patches (96 × 96 pixels) extracted from H&E-stained WSIs of sentinel lymph node sections in the CAMELYON16 dataset. Patches are binary-labeled for metastatic tissue presence, enabling rotation-equivariant CNN evaluation in digital pathology.

**MHIST** consists of 3,152 H&E-stained FFPE fixed-size images (224 × 224 pixels) of colorectal polyps from Dartmouth-Hitchcock Medical Center. Images are binary-classified as hyperplastic polyp or sessile serrated adenoma, with labels via majority vote from seven gastrointestinal pathologists.

**NCTCRC** consists of 100,000 non-overlapping patches (224 × 224 pixels) from H&E-stained histological images of colorectal cancer and normal tissue from the National Center for Tumor Diseases, Heidelberg. Patches are classified into nine categories: adipose, background, debris, lymphocytes, mucus, smooth muscle, normal colon mucosa, cancer-associated stroma, and colorectal adenocarcinoma epithelium.

**PatchVQA** consists of 5,382 pathology image patches extracted from whole-slide images, paired with 6,335 multiple-choice questions annotated and validated by professional pathologists. Questions incorporate distractor options to mitigate shortcut learning, enabling robust benchmarking of large multimodal models for patch-level visual question answering in pathology.

**GLySAC** consists of 59 H&E-stained image tiles (1,000 × 1,000 pixels) from gastric adenocarcinoma WSIs at Korea University Guro Hospital, acquired at 40× magnification. Nuclei (totaling 197,889) are segmented and classified into six types: epithelial, connective, lymphocyte, plasma, neutrophil, and eosinophil/macrophage.

**CoNSeP** consists of 41 H&E-stained images (1,000 × 1,000 pixels) from colorectal tissue at University Hospitals Coventry and Warwickshire. Nuclei (24,319 total) are segmented and classified into seven phenotypes: non-dysplastic epithelial, dysplastic epithelial, inflammatory, connective/soft tissue, necrotic debris, glandular secretions, and blood cells.

**HEMIT** consists of 1,085 paired H&E and multiplex-immunohistochemistry (mIHC) image tiles (512 × 512 pixels) from breast cancer tissue at the University of Manchester, with cellular-level registration. Pairs target translation for markers including CD3, CD4, CD8, CD20, CD68, FoxP3, and pan-cytokeratin.

**SHIFT** consists of 104 paired H&E and immunofluorescence (IF) whole-slide images from prostate core biopsies at the University of Pittsburgh, focusing on tumor signatures. Pairs enable translation from histological to immunofluorescent staining for markers like pan-cytokeratin, p63, and AMACR.

## Author contributions

X.C.G., Z.Y.F., and Y.F.W. conceived the study and designed the experiments. Z.Y.F. collected the data for Training. Z.Y.F., F.Q.C., Y.J.Z., Z.Y.C., H.X.M., and X.L. performed model development for all downstream tasks and the pathologist evaluations. Y.W.W., H.P.W., M.Z., and H.T.S. performed quality control of all the experiment results. X.C.G., Y.F.W., Z.Z., and Y.B.Z. prepared the manuscript. All authors contributed to the writing. Z.Z. and Y.B.Z. supervised the research.

## Conflict-of-interest disclosure

The authors declare no competing interests.

## Data availability

All data used in this study are publicly accessible. PubMedVision (https://huggingface.co/datasets/FreedomIntelligence/PubMedVision); PMC-OA (https://pmc.ncbi.nlm.nih.gov/tools/openftlist/); PMC-VQA (https://github.com/xiaoman-zhang/PMC-VQA); Quilt-1M (https://quilt1m.github.io/); PathGen-1.6M (https://huggingface.co/datasets/jamessyx/PathGen); BACH (https://iciar2018-challenge.grand-challenge.org/Dataset/); BRACS (https://www.bracs.icar.cnr.it/); BreakHis (https://web.inf.ufpr.br/vri/databases/breast-cancer-histopathological-database-breakhis/); Lunghist (https://figshare.com/articles/dataset/LungHist700_A_Dataset_of_Histological_Images_for_Deep_Learning_in_Pulmonary_Pathology/25459174); ARCH (https://warwick.ac.uk/fac/cross_fac/tia/data/arch); PathMMU (https://huggingface.co/datasets/jamessyx/PathMMU); ColonPath (https://springernature.figshare.com/articles/dataset/ColonPath_Tumor_Tissue_Screening_in_Pathology_Patches/22302799); Pcam (https://github.com/basveeling/pcam); MHIST (https://bmirds.github.io/MHIST/); HMNIST (https://zenodo.org/records/53169); NCTCRC (https://zenodo.org/records/1214456); PatchVQA (https://github.com/superjamessyx/PathBench); GLySAC, CoNSeP (https://github.com/QuIIL/Sonnet?tab=readme-ov-file); HEMIT (https://github.com/BianChang/HEMIT-DATASET); SHIFT (https://gitlab.com/eburling/SHIFT).

## Code availability

The trained model and source codes can be accessed at https://github.com/fzyzhiyuan/CRAFTS.

# Figures

**Fig. 1 | Overview of CRAFTS. a**, Pre-training dataset consisting of approximately 1.2 million image-caption pairs sourced from medical books and educational videos. **b**, Fine-tuning dataset comprising around 1.6 million image-caption pairs from TCGA, annotated with information on 30 cancer types, along with corresponding image feature distributions. **c**, Schematic of the CRAFTS framework: The pre-training phase

learns the semantic alignment between text and images, while the fine-tuning phase incorporates cancer-specific prior knowledge, with hallucination mitigation strategies applied throughout both stages of training. **d**, Evaluation of synthetic images through a combination of objective metrics and expert pathologist assessments. **e**, CRAFTS' data augmentation performance significantly outperforms existing state-of-the-art methods in Classification, Retrieval, Self-Supervised Learning (SSL), and Visual Question Answering (VQA) tasks. **f**, Integration of CRAFTS with ControlNet enables controllable generation of pathological image structures and semantics based on specific features, such as nuclear segmentation masks or fluorescence images.

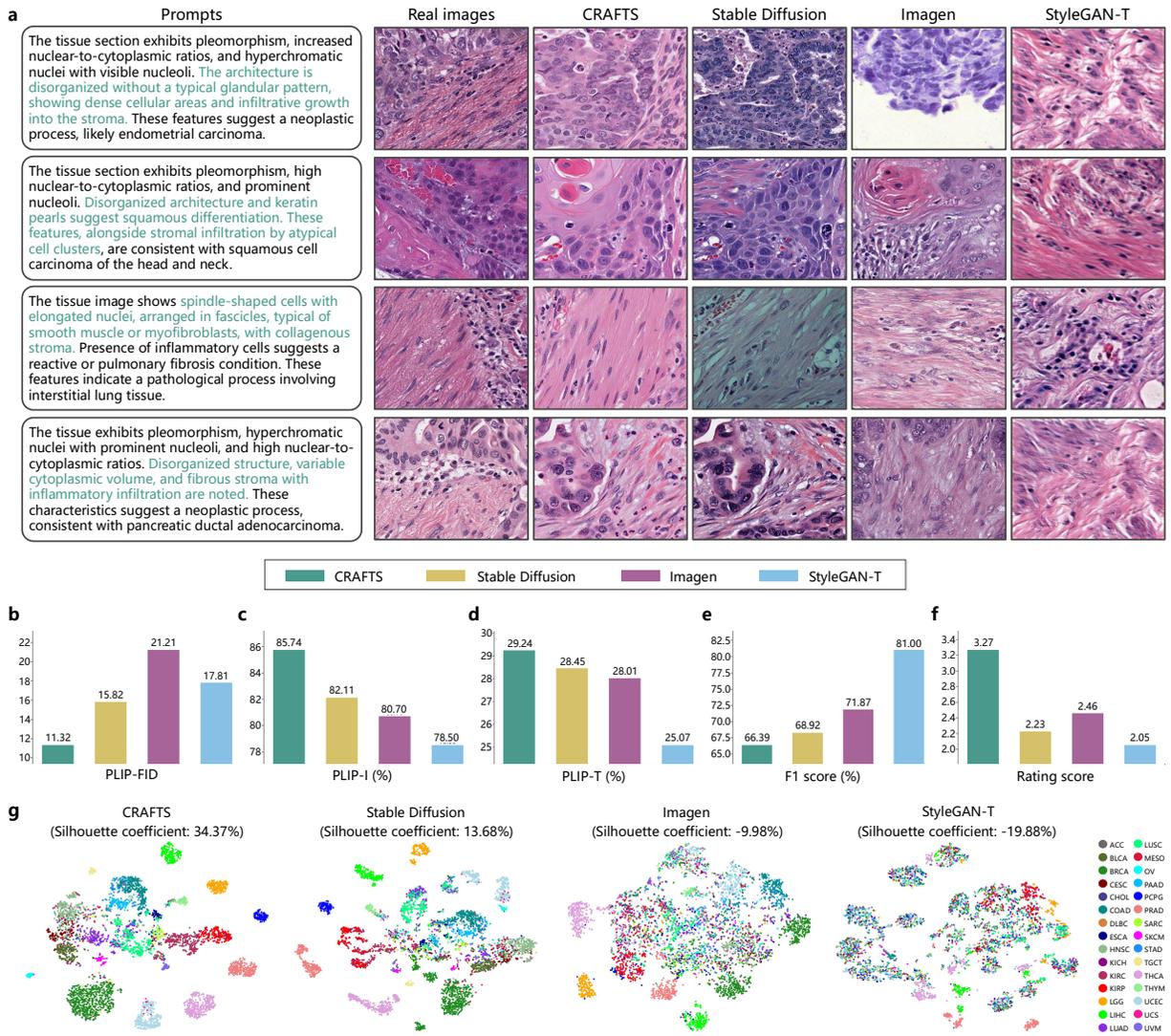

**Fig. 2 | Evaluation of synthetic images. a**, Example synthetic images generated by different methods. **b–d**, Performance comparison of CRAFTS and other state-of-the-art methods through objective evaluations. **b**, Diversity of synthetic pathological images (PLIP-FID). **c**, Semantic similarity between synthetic images and real images (PLIP-I). **d**, Semantic matching between synthetic images and text descriptions (PLIP-T). **e-f**, Pathologist evaluations of the synthetic images. **e**, F1 score for pathologists distinguishing synthetic images from real images. **f**, Pathologist ranking scores for the semantic alignment between synthetic images and text descriptions for CRAFTS and other state-of-the-art models. **g**, Distribution of class features for synthetic images generated by CRAFTS and other methods.

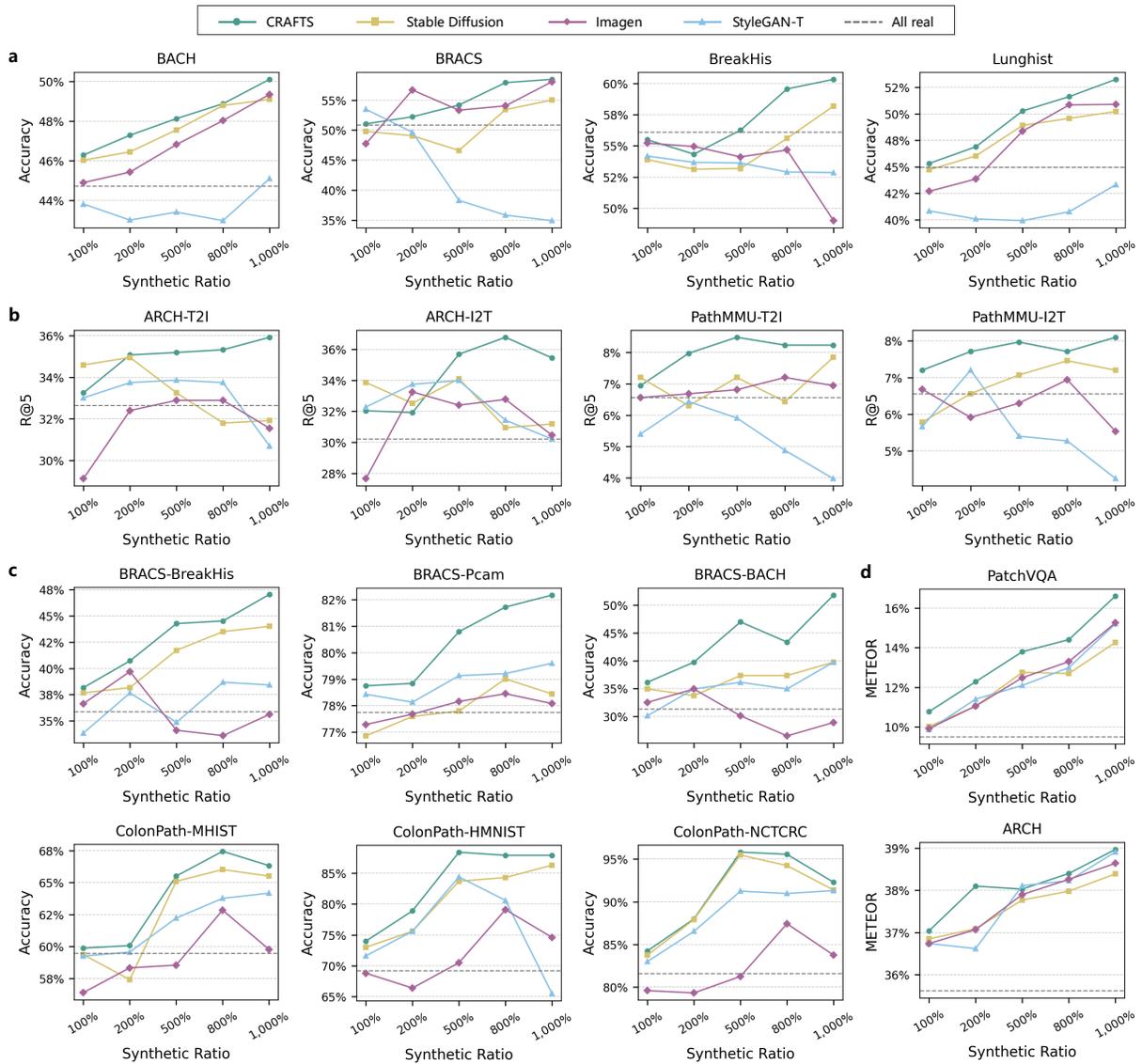

**Fig. 3 | Performance comparison of synthetic images from CRAFTS and other generative methods as data augmentation. a**, Performance comparison on image classification tasks across the BACH, BRACS, BreakHis, and Lunghist datasets. **b**, Performance comparison on cross-modal retrieval tasks across the ARCH and PathMMU datasets. **c**, Performance comparison on self-supervised learning tasks across the BRACS and ColonPath datasets. **d**, Performance comparison on visual question answering tasks across the PatchVQA and ARCH datasets.

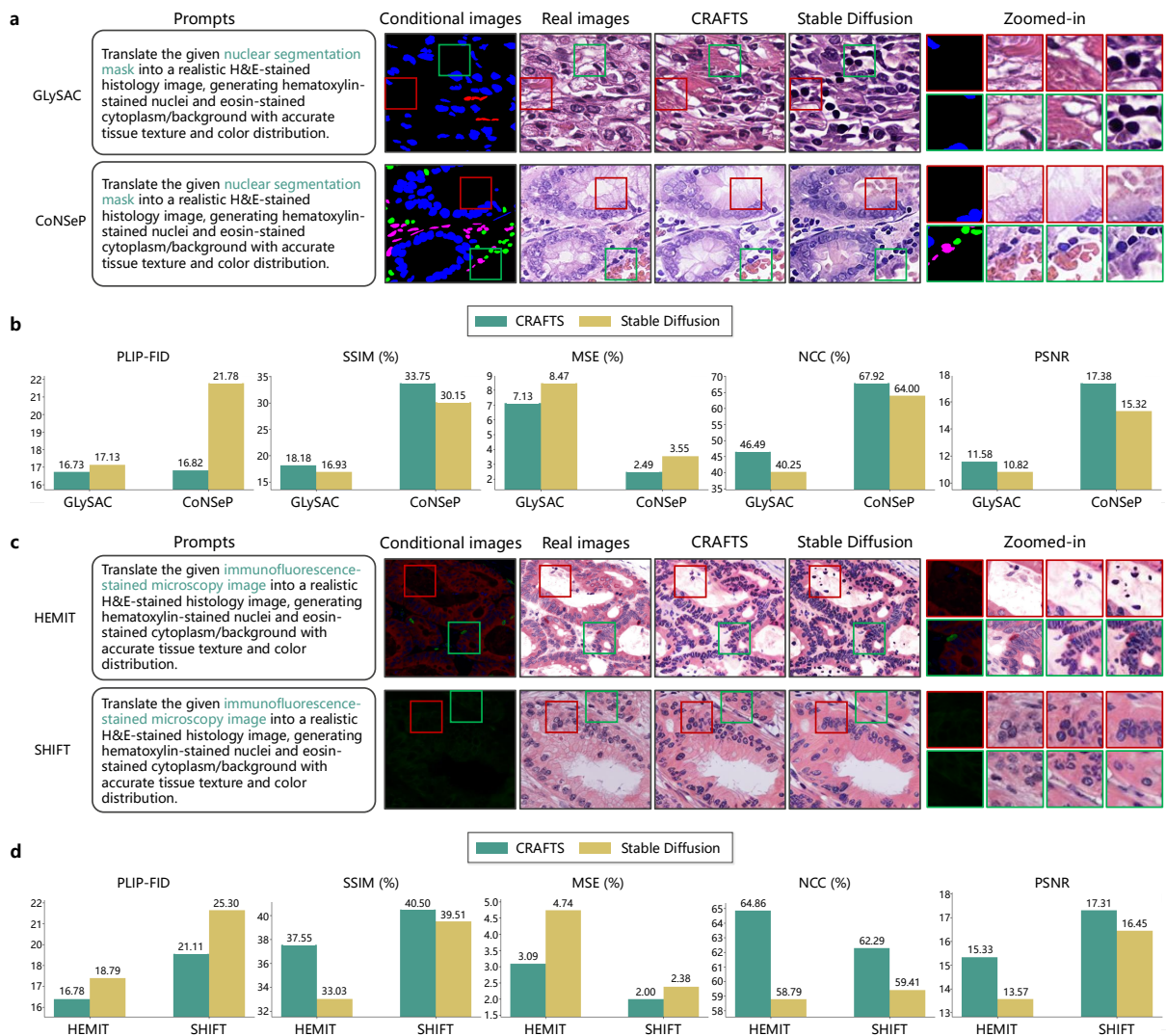

**Fig. 4 | Comparison of synthetic pathological images using CRAFTS and Stable Diffusion integrated with ControlNet, respectively. a**, Example synthetic images generated with segmentation masks as conditions on the GLySAC and CoNSeP datasets. **b**, Objective performance comparison of CRAFTS and Stable Diffusion on the GLySAC and CoNSeP datasets. **c**, Example synthetic images generated with fluorescence images as conditions on the HEMIT and SHIFT datasets. **d**, Objective performance comparison of CRAFTS and Stable Diffusion on the HEMIT and SHIFT datasets.

# Supplementary Information

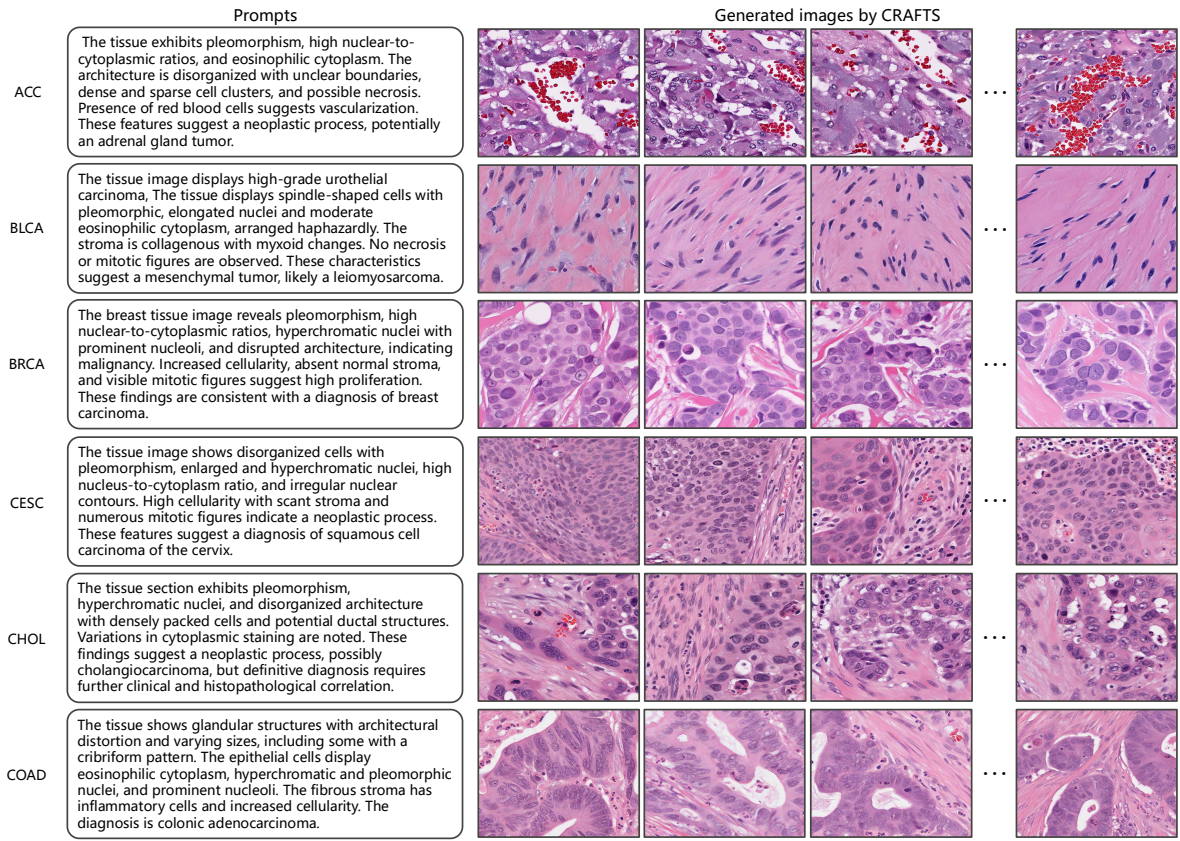

**Extended Data Fig. 1 | Examples of generated images by CRAFTS.** The first to sixth rows are Adrenocortical carcinoma (ACC), Bladder urothelial carcinoma (BLCA), Breast invasive carcinoma (BRCA), Cervical squamous cell carcinoma and endocervical adenocarcinoma (CESC), Cholangiocarcinoma (CHOL), Colon adenocarcinoma (COAD), respectively.

**Extended Data Fig. 2 | Examples of generated images by CRAFTS.** The first to sixth rows are Lymphoid neoplasm diffuse large B-cell lymphoma (DLBC), Esophageal carcinoma (ESCA), Head and Neck squamous cell carcinoma (HNSC), Kidney chromophobe (KICH), Kidney renal clear cell carcinoma (KIRC), Kidney renal papillary cell carcinoma (KIRP), respectively.

| | Prompts | Generated images by CRAFTS |
|---|---|---|
| LGG | The brain tissue section stained with H&E shows standard neuronal morphology with large, round nuclei and prominent nucleoli. The abundant eosinophilic cytoplasm and perineuronal net-like structures suggest a cerebral cortex origin. No signs of disorganization or necrosis are evident. The tissue structure is intact and the cell morphology is normal. | |
| LIHC | The liver tissue image shows hepatocytes with eosinophilic cytoplasm, nuclear pleomorphism, hyperchromatic nuclei, and prominent nucleoli, suggesting atypia. Dense cell packing and preserved tissue architecture with sinusoidal spaces are evident. Features are consistent with a potential hepatocellular adenoma or carcinoma. | |
| LUAD | The lung tissue exhibits significant pleomorphism, hyperchromatic and enlarged nuclei, high nuclear-to-cytoplasmic ratios, and dense cellular clusters, suggesting disrupted architecture. Atypical mitotic figures indicate active cell division. Prominent nucleoli and stromal infiltration by abnormal cells suggest a malignant neoplasm, likely carcinoma. | |
| LUSC | The lung tissue shows disrupted alveolar septa and spaces filled with mixed inflammatory cells, suggestive of an inflammatory response. Pneumocytes appear necrotic or detached. Increased septal cellularity and interstitial thickening indicate edema or inflammation. These features are consistent with pneumonia, reflecting significant pathology within the lung. | |
| MESO | The tissue exhibits pleomorphism, nuclear atypia, and hyperchromatic, enlarged nuclei, suggesting malignant transformation. Dense, irregular gland-like structures and a fibrous stroma with collagen and inflammatory cells are present, indicating a neoplastic process. The overall features are consistent with a potential carcinoma diagnosis. | |
| OV | The tissue shows a high nuclear-to-cytoplasmic ratio, pleomorphism, hyperchromatic and variable nuclei, prominent nucleoli, and disorganized architecture without normal structures. Mitotic figures indicate active proliferation. These features suggest a neoplastic process, consistent with ovarian carcinoma, reflecting aggressive cellular behavior and possible malignancy. | |

**Extended Data Fig. 3 | Examples of generated images by CRAFTS.** The first to sixth rows are Brain lower-grade glioma (LGG), Liver hepatocellular carcinoma (LIHC), Lung adenocarcinoma (LUAD), Lung squamous cell carcinoma (LUSC), Mesothelioma (MESO), Ovarian serous cystadenocarcinoma (OV), respectively.

| | Prompts | Generated images by CRAFTS | | | |
|---|---|---|---|---|---|
| PAAD | The tissue section shows spindle-shaped cells with elongated nuclei in a dense collagenous stroma, typical of connective tissue. Mild nuclear pleomorphism is present without necrosis or inflammation. These features suggest a spindle cell neoplasm of the pancreas, potentially a neuroendocrine tumor or sarcoma. Diagnosis requires further clinical evaluation. | 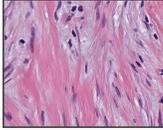 | 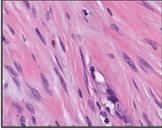 | 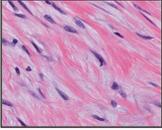 | ... 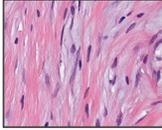 |
| PCPG | The tissue image exhibits high cellular density, disorganized arrangement, nuclear pleomorphism, and hyperchromatic nuclei, with eosinophilic cytoplasm. Cluster formation and red blood cells are present, suggesting altered adrenal gland structure. Further examination is needed to confirm the diagnosis of an adrenal gland tumor. | 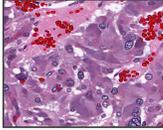 | 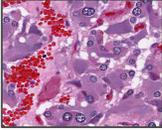 | 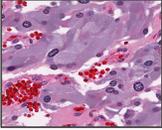 | ... 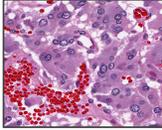 |
| PRAD | The microscopic image reveals prostate tissue with glandular hyperplasia, characterized by increased cells with round to oval nuclei and clear cytoplasm typical of clear cell cribriform hyperplasia. The luminal spaces are maintained, and the surrounding stroma is dense and fibrous. No signs of malignancy are present, with the glandular architecture preserved. | 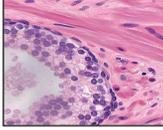 | 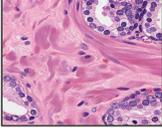 | 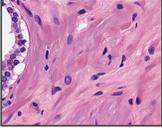 | ... 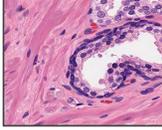 |
| SARC | The microscopic image reveals a mesenchymal neoplasm characterized by dense proliferation of uniform, spindle-shaped cells with elongated nuclei and moderate eosinophilic cytoplasm. These cells are disorganized and densely packed in areas, with scant extracellular matrix and interspersed collagen. Scattered red blood cells suggest vascular presence. | 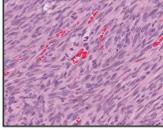 | 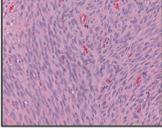 | 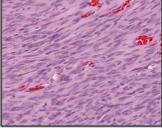 | ... 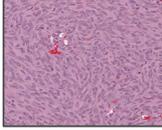 |
| SKCM | The skin tissue image shows a normal epidermis and dermis with fibroblasts, collagen fibers, and vascular structures. Clusters of inflammatory cells suggest a possible benign inflammatory response. Morphology supports a benign skin lesion, likely a dermatofibroma, characterized by fibroblast and histiocyte proliferation with inflammation. | 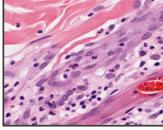 | 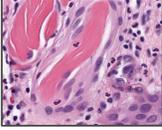 | 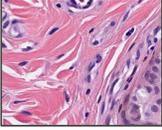 | ... 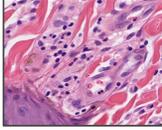 |
| STAD | The tissue image reveals a gastric mucosal section with preserved architecture. Features include a layer of columnar epithelial cells, dense lamina propria with potential inflammatory cells, and intact gastric glands lined by parietal or chief cells. There is increased cellularity in the lamina propria, suggesting inflammation. No signs of malignancy were observed. | 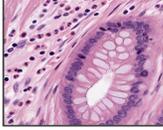 | 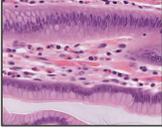 | 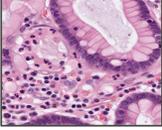 | ... 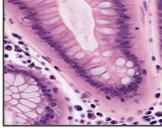 |

**Extended Data Fig. 4 | Examples of generated images by CRAFTS.** The first to sixth rows are Pancreatic adenocarcinoma (PAAD), Pheochromocytoma and paraganglioma (PCPG), Prostate adenocarcinoma (PRAD), Sarcoma (SARC), Skin cutaneous melanoma (SKCM), Stomach adenocarcinoma (STAD), respectively.

| | Prompts | Generated images by CRAFTS | | | | |
|---|---|---|---|---|---|---|
| TGCT | The microscopic image reveals densely packed cells with high pleomorphism, large hyperchromatic nuclei, and prominent nucleoli, indicative of neoplastic activity. The cells display variable eosinophilic and clear cytoplasm and lack normal testicular architecture. These characteristics suggest a malignant testicular germ cell tumor. | | | | ... | |
| THCA | The thyroid tissue image displays pleomorphism, varying cell sizes and shapes, enlarged overlapping irregular nuclei with prominent nucleoli, and disrupted follicular architecture. The colloid is reduced or absent. These findings suggest a neoplastic process, specifically papillary thyroid carcinoma, characterized by loss of normal tissue order and cellular atypia. | | | | ... | |
| THYM | The image shows a dense lymphocytic infiltrate in the thymus medulla, characterized by small, round cells with dark nuclei. The infiltrate is encapsulated, suggesting a lymphoid organ involvement. Such findings may indicate an inflammatory or autoimmune process, potentially involving thymitis. This could be associated with conditions like myasthenia gravis. | | | | ... | |
| UCEC | The tissue shows irregular, glandular structures with columnar epithelial cells exhibiting nuclear atypia, enlarged, hyperchromatic nuclei, indicative of malignancy. The disrupted architecture and presence of spindle-shaped stromal cells suggest a sarcomatous component. These findings are consistent with a diagnosis of mix of adenocarcinoma and sarcoma in the endometrium. | | | | ... | |
| UCS | The microscopic image reveals a dense cellular infiltrate with pleomorphic cells of varied sizes and shapes, hyperchromatic nuclei, high nuclear-to-cytoplasmic ratios, and prominent nucleoli. The fibrous stroma contains collagen. The disrupted tissue architecture suggests a neoplastic process, likely indicating a malignant uterine carcinoma. | | | | ... | |
| UVM | The tissue exhibits pleomorphism, hyperchromatic nuclei, irregular nuclear contours, variable cytoplasms, and high cellularity, indicative of a neoplastic process. Dense cellularity and scant stroma suggest invasiveness. Eosinophilic material and hemorrhage hint at stromal involvement. These features suggest a malignant tumor, likely uveal melanoma. | | | | ... | |

**Extended Data Fig. 5 | Examples of generated images by CRAFTS.** The first to sixth rows are Testicular germ cell tumors (TGCT), Thyroid carcinoma (THCA), Thymoma (THYM), Uterine corpus endometrial carcinoma (UCEC), Uterine carcinosarcoma (UCS), Uveal melanoma (UVM), respectively.

**Extended Data Tab. 1 | Comparison of performance between synthetic images generated by CRAFTS and other generative methods for data augmentation in image classification tasks on the BACH dataset.** The best result is marked in **bold**, and the second-best is underlined.

| Method | Synthetic ratio | Accuracy (%) | F1 score (%) | AUC (%) |
|---|---|---|---|---|
| All real | 0 | 44.73 | 44.64 | 70.29 |
| StyleGAN-T | 100% | 43.82 | 43.70 | 70.02 |
|  | 200% | 43.01 | 43.06 | 69.37 |
|  | 500% | 43.41 | 42.86 | 68.68 |
|  | 800% | 42.98 | 42.09 | 68.69 |
|  | 1,000% | 45.10 | 44.51 | 70.22 |
| Imagen | 100% | 44.90 | 44.48 | 70.05 |
|  | 200% | 45.43 | 45.01 | 70.86 |
|  | 500% | 46.82 | 46.46 | 72.04 |
|  | 800% | 48.03 | 47.56 | 73.09 |
|  | 1,000% | <u>49.35</u> | <u>49.30</u> | 73.59 |
| Stable Diffusion | 100% | 46.03 | 45.35 | 71.54 |
|  | 200% | 46.45 | 46.21 | 71.84 |
|  | 500% | 47.56 | 47.01 | 73.21 |
|  | 800% | 48.79 | 48.37 | 74.11 |
|  | 1,000% | 49.10 | 48.33 | <u>75.19</u> |
| CRAFTS | 100% | 46.29 | 46.33 | 71.63 |
|  | 200% | 47.29 | 46.54 | 71.93 |
|  | 500% | 48.13 | 47.56 | 73.38 |
|  | 800% | 48.88 | 48.36 | 74.15 |
|  | 1,000% | **50.11** | **49.59** | **75.57** |

**Extended Data Tab. 2 | Comparison of performance between synthetic images generated by CRAFTS and other generative methods for data augmentation in image classification tasks on the BRACS dataset.** The best result is marked in **bold**, and the second-best is underlined.

| Method | Synthetic ratio | Accuracy (%) | F1 score (%) | AUC (%) |
|---|---|---|---|---|
| All real | 0 | 50.88 | 52.07 | 66.75 |
| StyleGAN-T | 100% | 53.48 | 54.92 | 67.46 |
| | 200% | 49.65 | 50.49 | 62.72 |
| | 500% | 38.32 | 38.83 | 57.22 |
| | 800% | 35.87 | 36.31 | 56.99 |
| | 1,000% | 34.95 | 36.09 | 52.36 |
| Imagen | 100% | 47.73 | 50.06 | 66.32 |
| | 200% | 56.68 | 57.90 | 70.42 |
| | 500% | 53.33 | 54.39 | 67.15 |
| | 800% | 54.05 | 55.59 | 69.85 |
| | 1,000% | <u>58.06</u> | 58.70 | 69.77 |
| Stable Diffusion | 100% | 49.80 | 52.31 | 65.83 |
| | 200% | 49.08 | 50.96 | 64.03 |
| | 500% | 46.63 | 48.01 | 64.96 |
| | 800% | 53.42 | 55.14 | 69.21 |
| | 1,000% | 55.02 | 55.57 | 67.41 |
| CRAFTS | 100% | 51.06 | 52.14 | 66.65 |
| | 200% | 52.23 | 53.41 | 67.77 |
| | 500% | 54.19 | 54.60 | 68.53 |
| | 800% | 57.91 | <u>58.80</u> | <u>71.13</u> |
| | 1,000% | **58.46** | **59.38** | **72.43** |

**Extended Data Tab. 3 | Comparison of performance between synthetic images generated by CRAFTS and other generative methods for data augmentation in image classification tasks on the BreakHis dataset.** The best result is marked in **bold** and, the second-best is underlined.

| Method | Synthetic ratio | Accuracy (%) | F1 score (%) | AUC (%) |
| --- | --- | --- | --- | --- |
| All real | 0 | 56.10 | 52.68 | 81.53 |
| StyleGAN-T | 100% | 54.18 | 50.87 | 79.42 |
|  | 200% | 53.68 | 49.95 | 77.09 |
|  | 500% | 53.63 | 50.46 | 76.52 |
|  | 800% | 52.91 | 50.50 | 78.27 |
|  | 1,000% | 52.86 | 49.42 | 76.03 |
| Imagen | 100% | 55.22 | 53.15 | 82.73 |
|  | 200% | 54.95 | 52.99 | 82.98 |
|  | 500% | 54.12 | 53.32 | 83.51 |
|  | 800% | 54.67 | 53.06 | 81.82 |
|  | 1,000% | 49.01 | 47.69 | 79.43 |
| Stable Diffusion | 100% | 53.90 | 50.50 | 79.52 |
|  | 200% | 53.31 | 51.35 | 80.51 |
|  | 500% | 53.19 | 51.69 | 81.71 |
|  | 800% | 55.60 | 52.61 | 81.75 |
|  | 1,000% | 58.19 | 56.20 | 83.99 |
| CRAFTS | 100% | 55.49 | 53.07 | 82.30 |
|  | 200% | 54.34 | 52.73 | 82.09 |
|  | 500% | 56.26 | 55.16 | 83.87 |
|  | 800% | <u>59.56</u> | **59.00** | **86.82** |
|  | 1,000% | **60.33** | <u>58.89</u> | <u>85.46</u> |

**Extended Data Tab. 4 | Comparison of performance between synthetic images generated by CRAFTS and other generative methods for data augmentation in image classification tasks on the Lunghist dataset.** The best result is marked in **bold**, and the second-best is underlined.

| Method | Synthetic ratio | Accuracy (%) | F1 score (%) | AUC (%) |
|---|---|---|---|---|
| All real | 0 | 44.96 | 44.50 | 77.53 |
| StyleGAN-T | 100% | 40.84 | 40.36 | 74.87 |
| | 200% | 40.08 | 39.81 | 74.53 |
| | 500% | 39.92 | 39.74 | 74.31 |
| | 800% | 40.76 | 39.95 | 75.01 |
| | 1,000% | 43.32 | 42.35 | 76.97 |
| Imagen | 100% | 42.69 | 40.94 | 76.36 |
| | 200% | 43.85 | 43.68 | 77.05 |
| | 500% | 48.36 | 46.69 | 80.73 |
| | 800% | 50.84 | 50.41 | 82.54 |
| | 1,000% | 50.89 | 50.09 | 82.69 |
| Stable Diffusion | 100% | 44.72 | 44.04 | 78.07 |
| | 200% | 46.03 | 45.35 | 78.73 |
| | 500% | 48.91 | 48.58 | 81.04 |
| | 800% | 49.57 | 49.26 | 81.63 |
| | 1,000% | 50.22 | 50.02 | 81.98 |
| CRAFTS | 100% | 45.32 | 44.72 | 78.54 |
| | 200% | 46.89 | 46.09 | 79.67 |
| | 500% | 50.27 | 49.54 | 81.87 |
| | 800% | <u>51.62</u> | <u>50.98</u> | <u>82.77</u> |
| | 1,000% | **53.25** | **52.66** | **83.59** |

**Extended Data Tab. 5 | Comparison of performance between synthetic images generated by CRAFTS and other generative methods for data augmentation in text-to-image retrieval tasks on the ARCH dataset.** The best result is marked in **bold**, and the second-best is underlined.

| Method | Ratio | R@1 (%) | R@5 (%) | R@10 (%) | R@50 (%) |
|---|---|---|---|---|---|
| All real | 0 | 11.41 | 32.65 | 41.38 | 74.39 |
| StyleGAN-T | 100% | 11.29 | 33.01 | 42.96 | 77.43 |
| | 200% | 12.38 | 33.74 | 44.90 | 78.16 |
| | 500% | 13.11 | 33.86 | 46.60 | 76.70 |
| | 800% | 12.99 | 33.74 | 45.63 | 78.88 |
| | 1,000% | 11.41 | 30.70 | 43.20 | 76.58 |
| Imagen | 100% | 9.47 | 29.13 | 41.26 | 75.49 |
| | 200% | 13.59 | 32.40 | 46.97 | 77.67 |
| | 500% | 10.44 | 32.89 | 46.24 | 79.13 |
| | 800% | 12.50 | 32.89 | 46.72 | 78.64 |
| | 1,000% | 10.44 | 31.55 | 44.66 | 77.18 |
| Stable Diffusion | 100% | 12.86 | 34.59 | 45.39 | 75.00 |
| | 200% | 11.65 | 34.95 | 47.82 | 76.94 |
| | 500% | 12.62 | 33.25 | <u>48.91</u> | 78.76 |
| | 800% | 13.35 | 31.80 | 42.72 | 77.91 |
| | 1,000% | 11.17 | 31.92 | 44.54 | 76.94 |
| CRAFTS | 100% | 13.59 | 33.25 | 44.05 | 74.88 |
| | 200% | 12.62 | 35.07 | 46.00 | 78.28 |
| | 500% | **14.20** | 35.19 | 48.21 | <u>79.98</u> |
| | 800% | 13.59 | <u>35.32</u> | 48.54 | 79.13 |
| | 1,000% | <u>13.83</u> | **35.92** | **49.64** | **82.04** |

**Extended Data Tab. 6 | Comparison of performance between synthetic images generated by CRAFTS and other generative methods for data augmentation in image-to-text retrieval tasks on the ARCH dataset.** The best result is marked in **bold**, and the second-best is underlined.

| Method | Ratio | R@1 (%) | R@5 (%) | R@10 (%) | R@50 (%) |
|---|---|---|---|---|---|
| All real | 0 | 10.92 | 30.22 | 44.17 | 74.64 |
| StyleGAN-T | 100% | 11.77 | 32.28 | 44.30 | 76.33 |
|  | 200% | 12.26 | 33.74 | 47.21 | 77.79 |
|  | 500% | 10.44 | 33.98 | 46.24 | 79.25 |
|  | 800% | 10.68 | 31.43 | 45.63 | 78.52 |
|  | 1,000% | 11.17 | 30.22 | 41.87 | 76.94 |
| Imagen | 100% | 9.59 | 27.67 | 40.29 | 76.09 |
|  | 200% | 11.77 | 33.25 | 47.89 | 78.16 |
|  | 500% | 10.07 | 32.40 | 46.36 | 79.00 |
|  | 800% | 11.89 | 32.77 | 46.00 | 76.82 |
|  | 1,000% | 10.32 | 30.46 | 44.90 | 77.06 |
| Stable Diffusion | 100% | 12.38 | 33.86 | 45.15 | 75.97 |
|  | 200% | 11.29 | 32.52 | 45.75 | 76.46 |
|  | 500% | 12.50 | 34.10 | 48.18 | 78.03 |
|  | 800% | 10.80 | 30.95 | 44.90 | 75.97 |
|  | 1,000% | 12.01 | 31.19 | 46.60 | 77.79 |
| CRAFTS | 100% | 11.77 | 32.04 | 43.69 | 73.91 |
|  | 200% | 11.29 | 31.92 | 44.54 | 76.82 |
|  | 500% | <u>13.47</u> | <u>35.68</u> | 48.91 | <u>79.73</u> |
|  | 800% | **13.96** | **36.77** | <u>49.51</u> | 79.61 |
|  | 1,000% | 12.99 | 35.44 | **50.61** | **80.22** |

**Extended Data Tab. 7 | Comparison of performance between synthetic images generated by CRAFTS and other generative methods for data augmentation in text-to-image retrieval tasks on the PathMMU dataset.** The best result is marked in **bold**, and the second-best is underlined.

| Method | Ratio | R@1 (%) | R@5 (%) | R@10 (%) | R@50 (%) |
| --- | --- | --- | --- | --- | --- |
| All real | 0 | 1.54 | 6.56 | 11.70 | 29.18 |
| StyleGAN-T | 100% | 2.19 | 5.40 | 10.54 | 27.38 |
|  | 200% | 1.67 | 6.43 | 11.44 | 31.62 |
|  | 500% | 1.54 | 5.91 | 9.90 | 31.49 |
|  | 800% | 1.03 | 4.88 | 10.28 | 28.15 |
|  | 1,000% | 1.03 | 3.98 | 6.68 | 26.09 |
| Imagen | 100% | 1.80 | 6.56 | 11.57 | 28.53 |
|  | 200% | 1.93 | 6.68 | 11.18 | 30.72 |
|  | 500% | 2.06 | 6.81 | 11.05 | 31.62 |
|  | 800% | 1.93 | 7.20 | 12.21 | 31.88 |
|  | 1,000% | 1.54 | 6.94 | 11.95 | 29.95 |
| Stable Diffusion | 100% | 1.67 | 7.20 | 10.80 | 28.02 |
|  | 200% | 1.29 | 6.30 | 12.08 | 32.65 |
|  | 500% | 1.80 | 7.20 | 11.44 | 31.62 |
|  | 800% | 1.93 | 6.43 | 12.21 | 33.68 |
|  | 1,000% | <u>2.44</u> | 7.84 | 12.60 | 33.68 |
| CRAFTS | 100% | 1.93 | 6.94 | 11.31 | 30.21 |
|  | 200% | 2.06 | 7.97 | 12.08 | 32.13 |
|  | 500% | 2.19 | **8.48** | <u>12.98</u> | <u>35.09</u> |
|  | 800% | **2.57** | <u>8.23</u> | 11.44 | 33.55 |
|  | 1,000% | <u>2.44</u> | <u>8.23</u> | **13.24** | **35.86** |

**Extended Data Tab. 8 | Comparison of performance between synthetic images generated by CRAFTS and other generative methods for data augmentation in image-to-text retrieval tasks on the PathMMU dataset.** The best result is marked in **bold**, and the second-best is underlined.

| Method | Ratio | R@1 (%) | R@5 (%) | R@10 (%) | R@50 (%) |
|---|---|---|---|---|---|
| All real | 0 | 1.67 | 6.56 | 11.95 | 29.18 |
| StyleGAN-T | 100% | 1.16 | 5.66 | 10.41 | 28.53 |
|  | 200% | 1.16 | 7.20 | 11.57 | 29.95 |
|  | 500% | 1.67 | 5.40 | 9.25 | 29.82 |
|  | 800% | 0.90 | 5.27 | 9.77 | 28.53 |
|  | 1,000% | 1.16 | 4.24 | 6.94 | 26.61 |
| Imagen | 100% | 1.67 | 6.68 | 10.54 | 29.05 |
|  | 200% | 1.16 | 5.91 | 10.03 | 29.18 |
|  | 500% | 2.31 | 6.30 | 11.14 | 31.62 |
|  | 800% | 2.06 | 6.94 | 11.95 | 30.98 |
|  | 1,000% | 1.67 | 5.53 | 10.15 | 30.85 |
| Stable Diffusion | 100% | 2.19 | 5.78 | 10.15 | 26.35 |
|  | 200% | 1.41 | 6.56 | 10.41 | 32.65 |
|  | 500% | 1.93 | 7.07 | 11.31 | 31.62 |
|  | 800% | 2.31 | 7.46 | 11.44 | 34.70 |
|  | 1,000% | 1.80 | 7.20 | <u>12.08</u> | 33.80 |
| CRAFTS | 100% | 2.31 | 7.20 | <u>12.08</u> | 29.56 |
|  | 200% | <u>2.44</u> | 7.71 | 11.44 | 32.39 |
|  | 500% | 2.19 | <u>7.97</u> | **12.85** | <u>36.12</u> |
|  | 800% | **2.57** | 7.71 | 10.93 | 33.29 |
|  | 1,000% | 1.93 | **8.10** | **12.85** | **36.89** |

**Extended Data Tab. 9 | Performance comparison between images synthesized by CRAFTS and other generative methods when used as data augmentation for self-supervised learning on the BRACS dataset, with downstream evaluation conducted on the BreakHis dataset.** The best result is marked in **bold**, and the second-best is underlined.

| Method | Synthetic ratio | Accuracy (%) | F1 score (%) | AUC (%) |
|---|---|---|---|---|
| All real | 0 | 35.88 | 24.13 | 65.01 |
| StyleGAN-T | 100% | 33.84 | 21.10 | 64.10 |
|  | 200% | 37.66 | 23.35 | 65.95 |
|  | 500% | 34.86 | 21.66 | 63.30 |
|  | 800% | 38.68 | 21.32 | 66.04 |
|  | 1,000% | 38.42 | 20.81 | 69.09 |
| Imagen | 100% | 36.64 | 21.96 | 64.24 |
|  | 200% | 39.70 | 21.11 | 62.60 |
|  | 500% | 34.10 | 23.18 | 64.91 |
|  | 800% | 33.59 | 20.03 | 61.58 |
|  | 1,000% | 35.62 | 21.00 | 59.74 |
| Stable Diffusion | 100% | 37.66 | 24.29 | 68.75 |
|  | 200% | 38.17 | 24.00 | 69.24 |
|  | 500% | 41.73 | 25.68 | 71.88 |
|  | 800% | 43.51 | 24.81 | 73.32 |
|  | 1,000% | 44.02 | 27.97 | 74.69 |
| CRAFTS | 100% | 38.17 | 25.06 | 72.41 |
|  | 200% | 40.72 | 25.87 | 69.50 |
|  | 500% | 44.29 | 28.51 | 73.80 |
|  | 800% | <u>44.53</u> | <u>30.77</u> | <u>76.23</u> |
|  | 1,000% | **47.07** | **33.54** | **76.37** |

**Extended Data Tab. 10 | Performance comparison between images synthesized by CRAFTS and other generative methods when used as data augmentation for self-supervised learning on the BRACS dataset, with downstream evaluation conducted on the Pcam dataset.** The best result is marked in **bold**, and the second-best is underlined.

| Method | Synthetic ratio | Accuracy (%) | F1 score (%) | AUC (%) |
|---|---|---|---|---|
| All real | 0 | 77.75 | 77.68 | 86.41 |
| StyleGAN-T | 100% | 78.43 | 78.41 | 86.36 |
|  | 200% | 78.13 | 78.13 | 85.68 |
|  | 500% | 79.13 | 79.10 | 87.14 |
|  | 800% | 79.21 | 79.14 | 87.68 |
|  | 1,000% | 79.60 | 79.55 | 87.87 |
| Imagen | 100% | 77.29 | 77.27 | 85.36 |
|  | 200% | 77.68 | 77.67 | 85.48 |
|  | 500% | 78.16 | 78.15 | 85.98 |
|  | 800% | 78.45 | 78.45 | 85.76 |
|  | 1,000% | 78.08 | 78.06 | 86.04 |
| Stable Diffusion | 100% | 76.86 | 76.71 | 85.44 |
|  | 200% | 77.59 | 77.49 | 86.34 |
|  | 500% | 77.80 | 77.69 | 86.67 |
|  | 800% | 79.01 | 78.93 | 87.15 |
|  | 1,000% | 78.44 | 78.33 | 87.29 |
| CRAFTS | 100% | 78.75 | 78.68 | 86.96 |
|  | 200% | 78.84 | 78.79 | 87.09 |
|  | 500% | 80.79 | 80.75 | 88.88 |
|  | 800% | <u>81.72</u> | <u>81.69</u> | <u>89.72</u> |
|  | 1,000% | **82.17** | **82.15** | **90.30** |

**Extended Data Tab. 11 | Performance comparison between images synthesized by CRAFTS and other generative methods when used as data augmentation for self-supervised learning on the BRACS dataset, with downstream evaluation conducted on the BACH dataset.** The best result is marked in **bold**, and the second-best is underlined.

| Method | Synthetic ratio | Accuracy (%) | F1 score (%) | AUC (%) |
|---|---|---|---|---|
| All real | 0 | 31.33 | 29.88 | 58.47 |
| StyleGAN-T | 100% | 30.12 | 28.14 | 48.58 |
|  | 200% | 34.94 | 34.98 | 60.26 |
|  | 500% | 36.15 | 35.80 | 59.25 |
|  | 800% | 34.94 | 34.97 | 58.13 |
|  | 1,000% | 39.76 | 39.36 | 66.38 |
| Imagen | 100% | 32.53 | 31.25 | 55.51 |
|  | 200% | 34.94 | 34.81 | 58.28 |
|  | 500% | 30.12 | 22.07 | 50.01 |
|  | 800% | 26.51 | 24.48 | 49.53 |
|  | 1,000% | 28.92 | 26.66 | 54.94 |
| Stable Diffusion | 100% | 34.94 | 34.67 | 60.65 |
|  | 200% | 33.74 | 33.55 | 61.53 |
|  | 500% | 37.35 | 37.93 | 60.30 |
|  | 800% | 37.35 | 37.07 | 62.99 |
|  | 1,000% | 39.76 | 39.76 | 63.72 |
| CRAFTS | 100% | 36.15 | 36.64 | 62.62 |
|  | 200% | 39.76 | 40.20 | 62.84 |
|  | 500% | <u>46.99</u> | <u>45.66</u> | 64.10 |
|  | 800% | 43.37 | 42.54 | <u>67.07</u> |
|  | 1,000% | **51.81** | **51.90** | **73.63** |

**Extended Data Tab. 12 | Performance comparison between images synthesized by CRAFTS and other generative methods when used as data augmentation for self-supervised learning on the ColonPath dataset, with downstream evaluation conducted on the MHIST dataset.** The best result is marked in **bold**, and the second-best is underlined.

| Method | Synthetic ratio | Accuracy (%) | F1 score (%) | AUC (%) |
|---|---|---|---|---|
| All real | 0 | 59.47 | 54.77 | 57.26 |
| StyleGAN-T | 100% | 59.26 | 54.86 | 55.96 |
| | 200% | 59.57 | 54.29 | 55.99 |
| | 500% | 62.23 | 56.52 | 65.66 |
| | 800% | 63.77 | 53.50 | 68.28 |
| | 1,000% | 64.18 | 45.19 | 67.87 |
| Imagen | 100% | 56.40 | 51.19 | 53.36 |
| | 200% | 58.34 | 53.61 | 56.83 |
| | 500% | 58.55 | 52.36 | 58.36 |
| | 800% | 62.85 | 57.23 | 63.85 |
| | 1,000% | 59.78 | 50.97 | 58.03 |
| Stable Diffusion | 100% | 59.37 | 53.91 | 59.97 |
| | 200% | 57.42 | 51.03 | 55.35 |
| | 500% | 65.10 | **58.88** | 70.48 |
| | 800% | 66.02 | <u>58.09</u> | 71.87 |
| | 1,000% | 65.51 | 57.22 | 70.59 |
| CRAFTS | 100% | 59.88 | 52.77 | 59.10 |
| | 200% | 60.08 | 53.27 | 59.35 |
| | 500% | 65.51 | 56.32 | 71.32 |
| | 800% | **67.45** | 57.70 | **77.19** |
| | 1,000% | <u>66.33</u> | 56.52 | <u>75.97</u> |

**Extended Data Tab. 13 | Performance comparison between images synthesized by CRAFTS and other generative methods when used as data augmentation for self-supervised learning on the ColonPath dataset, with downstream evaluation conducted on the HMNIST dataset.** The best result is marked in **bold**, and the second-best is underlined.

| Method | Synthetic ratio | Accuracy (%) | F1 score (%) | AUC (%) |
|---|---|---|---|---|
| All real | 0 | 69.20 | 69.13 | 94.00 |
| StyleGAN-T | 100% | 71.60 | 71.42 | 94.43 |
|  | 200% | 75.60 | 75.39 | 95.30 |
|  | 500% | 84.40 | 84.20 | 98.05 |
|  | 800% | 80.60 | 80.32 | 96.97 |
|  | 1,000% | 65.50 | 64.24 | 90.99 |
| Imagen | 100% | 68.80 | 68.42 | 92.80 |
|  | 200% | 66.40 | 65.85 | 92.54 |
|  | 500% | 70.50 | 70.42 | 93.85 |
|  | 800% | 79.10 | 78.78 | 96.25 |
|  | 1,000% | 74.60 | 74.17 | 95.34 |
| Stable Diffusion | 100% | 73.00 | 72.66 | 95.09 |
|  | 200% | 75.60 | 75.31 | 95.37 |
|  | 500% | 83.70 | 83.68 | 97.92 |
|  | 800% | 84.30 | 84.24 | 98.15 |
|  | 1,000% | 86.30 | 86.19 | 98.58 |
| CRAFTS | 100% | 74.00 | 74.06 | 95.38 |
|  | 200% | 78.90 | 78.94 | 96.35 |
|  | 500% | **88.40** | **88.31** | 98.44 |
|  | 800% | <u>87.90</u> | 87.77 | <u>98.67</u> |
|  | 1,000% | <u>87.90</u> | 87.84 | **98.70** |

**Extended Data Tab. 14 | Performance comparison between images synthesized by CRAFTS and other generative methods when used as data augmentation for self-supervised learning on the ColonPath dataset, with downstream evaluation conducted on the NCTCRC dataset.** The best result is marked in **bold**, and the second-best is underlined.

| Method | Synthetic ratio | Accuracy (%) | F1 score (%) | AUC (%) |
|---|---|---|---|---|
| All real | 0 | 81.60 | 74.71 | 96.72 |
| StyleGAN-T | 100% | 82.99 | 76.03 | 97.50 |
|  | 200% | 86.53 | 80.71 | 98.45 |
|  | 500% | 91.25 | 88.15 | 99.42 |
|  | 800% | 90.97 | 87.88 | 99.05 |
|  | 1,000% | 91.32 | 88.22 | 99.31 |
| Imagen | 100% | 79.58 | 73.48 | 96.71 |
|  | 200% | 79.31 | 73.44 | 95.99 |
|  | 500% | 81.25 | 75.43 | 96.60 |
|  | 800% | 87.43 | 82.57 | 98.47 |
|  | 1,000% | 83.75 | 76.50 | 97.50 |
| Stable Diffusion | 100% | 83.75 | 78.96 | 97.78 |
|  | 200% | 87.92 | 84.13 | 98.78 |
|  | 500% | 95.49 | <u>94.16</u> | 99.78 |
|  | 800% | 94.24 | 92.64 | 99.72 |
|  | 1,000% | 91.39 | 87.66 | 99.35 |
| CRAFTS | 100% | 84.24 | 79.58 | 98.11 |
|  | 200% | 87.99 | 84.20 | 98.92 |
|  | 500% | **95.83** | **94.35** | <u>99.80</u> |
|  | 800% | <u>95.57</u> | 93.74 | **99.82** |
|  | 1,000% | 92.29 | 89.09 | 99.63 |

**Extended Data Tab. 15 | Comparison of performance between synthetic images generated by CRAFTS and other generative methods for data augmentation in visual question answering tasks on the PatchVQA dataset.** The best result is marked in **bold**, and the second best is underlined.

| Method | Synthetic ratio | BLEU1 (%) | METEOR (%) | CIDER (%) |
|---|---|---|---|---|
| All real | 0 | 10.67 | 9.50 | 31.88 |
| StyleGAN-T | 100% | 11.09 | 9.86 | 35.50 |
|  | 200% | 12.98 | 11.40 | 41.87 |
|  | 500% | 13.67 | 12.09 | 46.41 |
|  | 800% | 14.26 | 12.99 | 51.02 |
|  | 1,000% | 17.22 | 15.20 | <u>63.23</u> |
| Imagen | 100% | 11.14 | 9.92 | 34.16 |
|  | 200% | 12.76 | 11.05 | 40.17 |
|  | 500% | 13.68 | 12.48 | 44.57 |
|  | 800% | 14.94 | 13.29 | 50.53 |
|  | 1,000% | <u>17.30</u> | <u>15.25</u> | 61.75 |
| Stable Diffusion | 100% | 11.67 | 10.02 | 34.88 |
|  | 200% | 12.86 | 11.05 | 38.90 |
|  | 500% | 14.63 | 12.75 | 50.43 |
|  | 800% | 14.16 | 12.69 | 51.25 |
|  | 1,000% | 15.99 | 14.26 | 58.14 |
| CRAFTS | 100% | 12.25 | 10.77 | 37.31 |
|  | 200% | 13.91 | 12.28 | 48.18 |
|  | 500% | 15.52 | 13.79 | 54.55 |
|  | 800% | 16.28 | 14.40 | 55.92 |
|  | 1,000% | **18.88** | **16.60** | **68.48** |

**Extended Data Tab. 16 | Comparison of performance between synthetic images generated by CRAFTS and other generative methods for data augmentation in visual question answering tasks on the ARCH dataset.** The best result is marked in **bold**, and the second best is underlined.

| Method | Synthetic ratio | BLEU1 (%) | METEOR (%) | CIDER (%) |
|---|---|---|---|---|
| All real | 0 | 35.00 | 35.62 | 89.64 |
| StyleGAN-T | 100% | 36.76 | 36.74 | 91.67 |
| | 200% | 36.72 | 36.62 | 94.07 |
| | 500% | 38.14 | 38.11 | 99.48 |
| | 800% | 38.43 | 38.23 | 99.74 |
| | 1,000% | <u>39.08</u> | <u>38.91</u> | <u>102.35</u> |
| Imagen | 100% | 36.50 | 36.74 | 93.58 |
| | 200% | 36.74 | 37.07 | 91.04 |
| | 500% | 38.21 | 37.90 | 98.49 |
| | 800% | 38.58 | 38.26 | 97.86 |
| | 1,000% | 39.03 | 38.64 | 101.31 |
| Stable Diffusion | 100% | 36.65 | 36.85 | 93.70 |
| | 200% | 36.89 | 37.09 | 94.05 |
| | 500% | 37.82 | 37.77 | 98.19 |
| | 800% | 37.96 | 37.98 | 97.77 |
| | 1,000% | 38.61 | 38.39 | 99.32 |
| CRAFTS | 100% | 37.13 | 37.04 | 95.88 |
| | 200% | 38.21 | 38.10 | 98.02 |
| | 500% | 38.35 | 38.03 | 99.62 |
| | 800% | 38.49 | 38.40 | 100.85 |
| | 1,000% | **39.27** | **38.97** | **103.40** |